%% file: main.tex
\documentclass[sigconf,9pt,nonacm,anonymous=false]{acmart}
\usepackage{hyperref}
\usepackage{cleveref}
\usepackage{graphicx}
\usepackage{textcomp}
\usepackage{etoolbox}
\usepackage{paralist}
\usepackage{booktabs}
\usepackage{multirow}
\usepackage{xcolor}
\usepackage{algorithm}
\usepackage{algorithmicx}
\usepackage{algpseudocode}
\usepackage{caption}
\usepackage{subcaption}
\usepackage{hyperref}
\newcommand{\method}{\texttt{FedCode}}

\begin{document}
\pagestyle{empty}

\title{FedCode: Communication-Efficient Federated Learning via Transferring Codebooks}

\begin{abstract}
Federated Learning (FL) is a distributed machine learning paradigm that enables learning models from decentralized local data. While FL offers appealing properties for clients' data privacy, it imposes high communication burdens for exchanging model weights between a server and the clients. Existing approaches rely on model compression techniques, such as pruning and weight clustering to tackle this. However, transmitting the entire set of weight updates at each federated round, even in a compressed format, limits the potential for a substantial reduction in communication volume. We propose~\method~where clients transmit only codebooks, i.e., the cluster centers of updated model weight values. To ensure a smooth learning curve and proper calibration of clusters between the server and the clients,~\method~ periodically transfers model weights after multiple rounds of solely communicating codebooks. This results in a significant reduction in communication volume between clients and the server in both directions, without imposing significant computational overhead on the clients or leading to major performance degradation of the models. We evaluate the effectiveness of~\method~using various publicly available datasets with ResNet-20 and MobileNet backbone model architectures. Our evaluations demonstrate a 12.2-fold data transmission reduction on average while maintaining a comparable model performance with an average accuracy loss of $1.3\%$ compared to \textit{FedAvg}. Further validation of~\method~performance under non-IID data distributions showcased an average accuracy loss of $2.0\%$ compared to \textit{FedAvg} while achieving approximately a 12.7-fold data transmission reduction. 
\end{abstract}
  

\author{Saeed Khalilian}
\affiliation{%
\institution{Eindhoven University of Technology}
\country{The Netherlands}}
\email{s.khalilian.gourtani.tue.nl}

\author{Vasileios Tsouvalas}
\affiliation{%
\institution{Eindhoven University of Technology}
\country{The Netherlands}}
\email{v.tsouvalas@tue.nl}

\author{Tanir Ozcelebi}
\affiliation{%
\institution{Eindhoven University of Technology}
\country{The Netherlands}}
\email{t.ozcelebi@tue.nl}

\author{Nirvana Meratnia}
\affiliation{%
\institution{Eindhoven University of Technology}
\country{The Netherlands}}
\email{n.meratnia@tue.nl}
\maketitle

\section{Introduction}\label{sec:introduction}

Federated Learning (FL) is an emerging learning paradigm that enables multiple devices to collaboratively train a deep neural network (DNN), while preserving data locally on these devices~\cite{fl}. In FL, clients' data remain local and only model updates are communicated to a central server, after which the global model at the server will be updated using an aggregation strategy (e.g., \textit{FedAvg}~\cite{fedavg}). Despite its appealing properties in terms of preserving users' privacy and mitigating the need to collect data in a central repository for learning, FL introduces communication overhead, as clients need to frequently communicate their model updates with the central server. This becomes even more critical when clients are resource-constrained edge devices with limited communication bandwidth and energy resources~\cite{OpenChallenges}. For these devices, it is also crucial that their computational limitations are taken into account, as some approaches impose a considerable computational overhead on the clients in order to reduce communication~\cite{fedzip, gtop_k, mucsc}.

To address the communication overhead in FL, different approaches such as improving training convergence to reduce the number of model updates to be communicated~\cite{oneshotFL, distilledoneshotFL, fedboost, fedadp, fednova, folb}, utilizing efficient compression techniques for message exchange~\cite{fedpaq, NIPS2017_6c340f25}, using one-shot learning~\cite{oneshotFL, distilledoneshotFL}, or increasing convergence rates by sampling participating devices~\cite{folb} can be found in the literature. Additionally, various server-side aggregation mechanisms have been proposed to achieve fast-convergence in FL and to reduce communication costs~\cite{fedboost, fedadp, fednova}. Recently, researchers explored model compression techniques such as sparsification~\cite{fedsparsify}, quantization~\cite{fedpaq}, and weight clustering~\cite{fedzip, mucsc} to achieve significant compression of model weight matrices before exchanging them between the clients and the server. For example, FedPAQ~\cite{fedpaq} quantizes the weight matrices to shorter bit representations for the purpose of communication-efficiency in FL, while FedZip~\cite{fedzip} combines sparsification with weight clustering in each model weight matrix (i.e., model layer) to reduce the number of unique elements in the weight matrices; thus, reducing the volume of exchanged messages.

\begin{figure*}[t]
    \centering
    \includegraphics[width=0.75\textwidth]{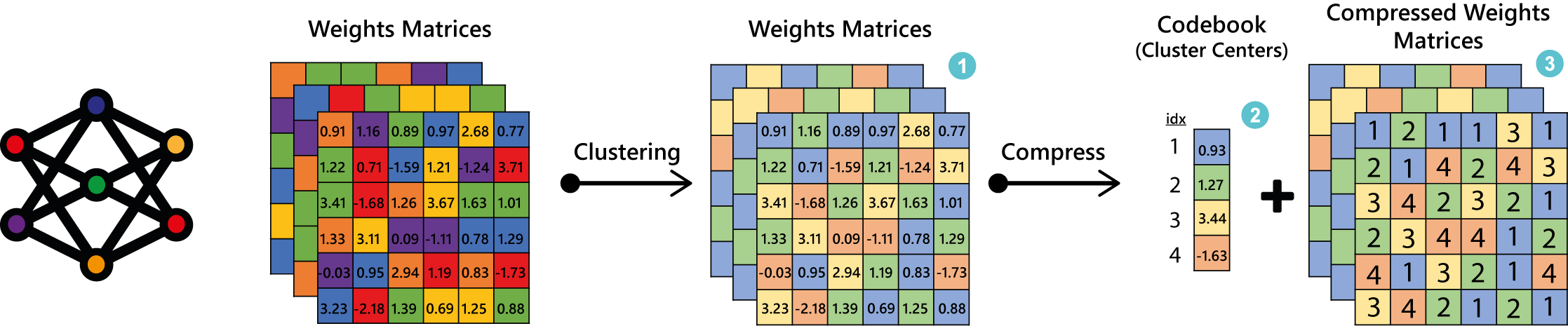}
    \caption{\small{Illustration of the weight clustering process, which involves: 1) clustering weight values in the weight matrices, 2) saving cluster centers in an array called the codebook, and 3) compressing the weight matrices by replacing each weight value with its corresponding cluster index.}\label{fig:weight_clustering}}
\end{figure*}

While prior approaches effectively reduce the model size, communicating all weight updates between the server and the clients at each federated round limits their potential for a substantial reduction of transmitted data volume. To minimize data transmission and to preserve bandwidth in the FL communication channel, we introduce~\method, which primarily transfers the cluster centers of weights (the so called codebooks) during the FL training process, unleashing the true potential of weight clustering and achieving even greater reduction in volume of exchanged messages in the bidirectional communication route. Furthermore,~\method~imposes minimal computational overhead for the clients; during the downstream communication route, clients perform a computationally inexpensive binary search for weight decompression from the received codebook. Moreover, uploading clients' updates to the server involves a single K-means clustering execution for the entire clients' model weights to extract the new codebook, after which the server will perform global model aggregation using the ``\textit{aggregated}'' codebooks. To ensure proper calibration of the clusters between the server and the clients, we periodically transfer the entire set of weights after clustering. By doing so, we achieve a significant reduction in data transmission while preserving global model accuracy. The main contributions of this paper are as follows:

\begin{itemize}
\item We propose a novel communication-efficient FL scheme, termed~\method\footnotemark[1], which reduces data transmission in the bidirectional communication channel by relying primarily on transferring clustered centers of weights (codebooks) instead of communicating entire weight matrices.
\item We present an effective ``\textit{codebook aggregation}'' mechanism on the server side to update global model weights while keeping the computational overhead on the clients low.
\item We evaluate the effectiveness of~\method~on vision and audio datasets, namely CIFAR-10~\cite{krizhevsky2009learning}, CIFAR-100~\cite{krizhevsky2009learning}, and SpeechCommands~\cite{spcm}, with two distinct backbone model architectures, i.e., MobileNet~\cite{sandler2018mobilenetv2} and ResNet-20~\cite{he2016deep}. 
\item Our evaluation shows, compared with \textit{FedAvg}, on average a 12.2-fold data transmission reduction with an average accuracy loss of 1.3\% can be achieved across all three datasets. Furthermore, our experiments on non-IID data distribution show that~\method~can achieve comparable performance to \textit{FedAvg}.
\end{itemize}


\section{Background}\label{sec:background}

In this section, we present the learning objectives of FL, as well as an overview of weight clustering and codebook creation processes for compressing deep learning models to provide the necessary foundations for our communication-efficient FL approach.

\subsection{Federated Learning}\label{ssec:fl}

FL~\cite{fl} is a collaborative learning paradigm that aims to learn a single global model from data stored on remote clients with no need to share their data with a central server. Specifically, with the data residing on clients' devices, a subset of clients is selected to perform a number of local model update steps on their data in parallel at each communication round. Upon completion, clients transmit their model weight updates to the server and receive the global model's weight updates, which aims to learn a unified global model by aggregating these updates. Formally, the goal of FL is to minimize the following objective function:

\begin{equation} \label{eqn:FL}
     \min_{\theta} \mathcal{L}_{\theta} = \sum_{m=1}^{M} \gamma_{m} {\mathcal{L}}_m(\theta),
\end{equation}

\noindent where $\mathcal{L}_m$ is the loss function of the $m^{th}$ client and $\gamma_{m}$ corresponds to the relative impact of the $m^{th}$ client on the construction of the global model. For the \textit{FedAvg}~\cite{fedavg} algorithm, $\gamma_{m}$ is equal to the ratio of client's local data $N_m$ over all training samples, i.e., $\left (\gamma_{m} = \frac{N_m}{N}\right)$. 

Here, the volume of exchanged updates in terms of bits downloaded and uploaded during FL training stage, is as follows:

\begin{equation}\label{eqn:fl_com}
    2 \times R \times P \times  worldlength,
\end{equation}

\noindent where $R$ refers to the number of federated rounds, $P$ is the number of trainable model parameters, and $\textit{worldlength}$ is the number of bits required to represent each weight element.

\begin{table*}[!t]
    \centering \small
   \caption{\small{Comparison of~\method~with state-of-the-art communication-efficient FL schemes. For client-side complexity, $l$ refers to the number of model layers, $p$ is the number of trainable model parameters, wand $p_l$ refers to the number of trainable parameters for a given layer $l$, $i$ depicts the number of iterations in the clustering algorithm; and $k$ represents the number of clusters in the clustering algorithm. \label{tab:related_work}}}
    \resizebox{0.75\hsize}{!}{%
        \begin{tabular}{lcccc}
            \toprule
            \textbf{Method}
            & \begin{tabular}[c]{c}\textbf{Compression}\\\textbf{Technique}\end{tabular}
            & \begin{tabular}[c]{c}\textbf{Compression}\\\textbf{Route}\end{tabular}
            & \begin{tabular}[c]{c}\textbf{Wide Compression}\\\textbf{Levels}\end{tabular}
            & \begin{tabular}[c]{c}\textbf{Client-side}\\\textbf{Complexity}\end{tabular} \\
            \midrule
            FedPAQ~\cite{fedpaq} & Quantization                 & Upstream                  &                               & $\mathcal{O}(p)$ \\
            FedSparsify~\cite{fedsparsify} & Threshold Pruning  & Bidirectional             &                               & $\mathcal{O}(p)$ \\
            gTop-k~\cite{gtop_k} & Top-k Pruning                & Bidirectional             & \checkmark                    & $\mathcal{O}(p\cdot logk)$ \\
            \midrule
            MUCSC~\cite{mucsc} & Layer-wise Clustering          & Bidirectional             &                               & $\mathcal{O}(l \cdot p_l\cdot i \cdot k )$ \\
            \multirow{2}{*}{FedZip~\cite{fedzip}} 
                        & Top-k Pruning \&                      & \multirow{2}{*}{Upstream} & \multirow{2}{*}{\checkmark}   & \multirow{2}{*}{$\mathcal{O}(l \cdot p_l\cdot i \cdot k )$} \\
                        & Layer-wise Clustering                 &                           &                               & \\
            FedCode & Weights Clustering                        & Bidirectional             & \checkmark                    & $\mathcal{O}(p \cdot i \cdot k )$ \\
            \bottomrule
        \end{tabular}%
        }
\end{table*}

\subsection{Weight Clustering and Codebook Creation}\label{ssec:wc}
The weight clustering process~\cite{han2015deep} encompasses three main steps, as depicted in Figure \ref{fig:weight_clustering}. First, weight values in weight matrices are grouped into clusters using a clustering algorithm like K-means~\cite{kmeans}. Second, the centers of these clusters are stored in an array known as the codebook. Finally, the weight values stored in weight matrices are replace with their corresponding cluster index. Weight clustering can be applied either on a per-layer basis or to the entire model. When applied to the entire model, the clustering algorithm is applied once to the entire set of weight values, whereas layer-wise weight clustering requires applying the clustering algorithm to each layer individually.

Formally, let us consider a network $p_{\theta}$ with weights $\theta$, where $P$ represents the number of trainable parameters and ``\textit{wordlength}'' denotes the number of bits required to represent each weight element. In this case, the model size (i.e., size of model weight matrices) is given by $|M|= P \times \textit{wordlength}$. Applying weight clustering in model $p_{\theta}$, the compressed model size is given by: 

\begin{equation}\label{eqn:dtr}
    |M'|=  K \times worldlength + P \times  \lceil \log_2 K \rceil,
\end{equation}

\noindent where $K$ is the number of clusters, $K \times \textit{worldlength}$ is the size of the codebook in bits, and $\lceil \log_2 K \rceil$ is the number of bits required to represent $K$ cluster indexes. Thus, the compressed model encompasses a codebook, storing cluster centers, and the compressed weight matrices, containing cluster indices.
 
\section{Related Work}
Communication reduction is one of the primary challenges in improving scalability of FL. Research on communication-efficient FL schemes generally revolves around two main areas: (i) reducing the number of communication rounds, and (ii) deploying compression schemes on the exchanged messages. In the former category, periodic communication is commonly used to reduce the amount of communicated messages. Clients perform multiple local training steps before transmitting their gradient updates to the server. Here, the reduction in communication correlates with the number of local training iterations, while more local iterations in between updates mean increased risk of misalignment in clients' training objectives, potentially degrading global model’s performance. Based on FedAvg~\cite{fedavg}, FedProx~\cite{fedprox} introduced a proximal term to the clients' local training objectives, effectively allowing for more local training steps without significant gradients divergence. Furthermore, FedNova~\cite{fednova} reduces communication by letting clients perform arbitrary number of local updates based on their resources and aggregating normalized gradient updates instead of weighted averaging. Apart from periodic communication, one-shot learning ~\cite{oneshotFL, distilledoneshotFL} has been explored to achieve fast convergence in FL. FOLB~\cite{folb} reduces communication rounds by intelligently sampling participating devices.

In communication compression, the focus shifts towards reducing the size of transmitted messages (i.e., weights matrices) at each federated round. Based on the idea that larger values contribute more to learning objectives, sparsification, also known as pruning, discards weights elements with small absolute values. Here, discarding elements can be done randomly~\cite{random_k}, using a magnitude threshold~\cite{fedsparsify}, or based on the largest absolute values~\cite{top_k, gtop_k}. It is possible to combine pruning with lossless compression algorithms, like Huffman Encoding~\cite{huffman1952method}. In gTop-k~\cite{gtop_k}, model pruning is introduced for both the clients' and the server's models, compressing model weight updates in both upstream and downstream communication route. FedSparsify~\cite{fedsparsify} utilizes magnitude pruning with a gradually increasing threshold during training to learn a highly-sparse model in a communication-efficient FL scheme. Taking a different approach, ProgFed~\cite{progfed} gradually increases the model capacity by training on shallow sub-models with fewer parameters, saving two-way communication costs in the early stages of the FL process.

Scalar quantization \cite{fedpaq,DAdaQuant} reduces communication overhead by reducing the precision of elements in the weight matrices (i.e., shorter bit precision). In FedPAQ~\cite{fedpaq}, authors combined periodic communication and scalar quantization for improved communication efficiency. DAdaQuant~\cite{DAdaQuant} utilizes a client-adaptive precision in the quantization algorithm to increase model accuracy, while maintaining low upstream communication cost. As opposed to quantizing matrix elements to shorter bit representations, quantization based on codebook mapping (referred to as weight clustering) converts the weight matrices elements into a discrete set of values; thus the number of unique elements of the weight matrices are quantized. This way, weight clustering achieves a higher compression ratio compared to scalar quantization, as it allows a small number of discrete values (codebook) to be transmitted together with their positions in the weight matrices (compressed weight matrices), whereas scalar quantization requires each element to consume at least one bit~\cite{bingham2022legonet, tanaka2020pruning}. MUCSC~\cite{mucsc} utilizes layer-wise weight clustering to communicate compressed weights between clients and server, while FedZip~\cite{fedzip} employs a sequence of Top-k~\cite{top_k} pruning, layer-wise weight clustering, and Huffman Encoding to compress the exchanged messages from clients to the server. One of the challenges in layer-wise weight clustering is determining the optimal number of clusters for each layer of the model. Various methods have been employed to explore the trade-off between compression ratio and model accuracy in weight clustering. Some approaches manually adjust the number of clusters, using either the same number of clusters for each layer, or different cluster counts for different layers~\cite{wu2018deep, son2018clustering, wu2016quantized}. Recently, heuristic algorithms~\cite{dupuis2022heuristic, dupuis2020automatic} have been explored to automatically determine the optimal number of shared values (clusters) per layer.

\begin{figure*}[!t]
    \centering
    \includegraphics[width=\textwidth]{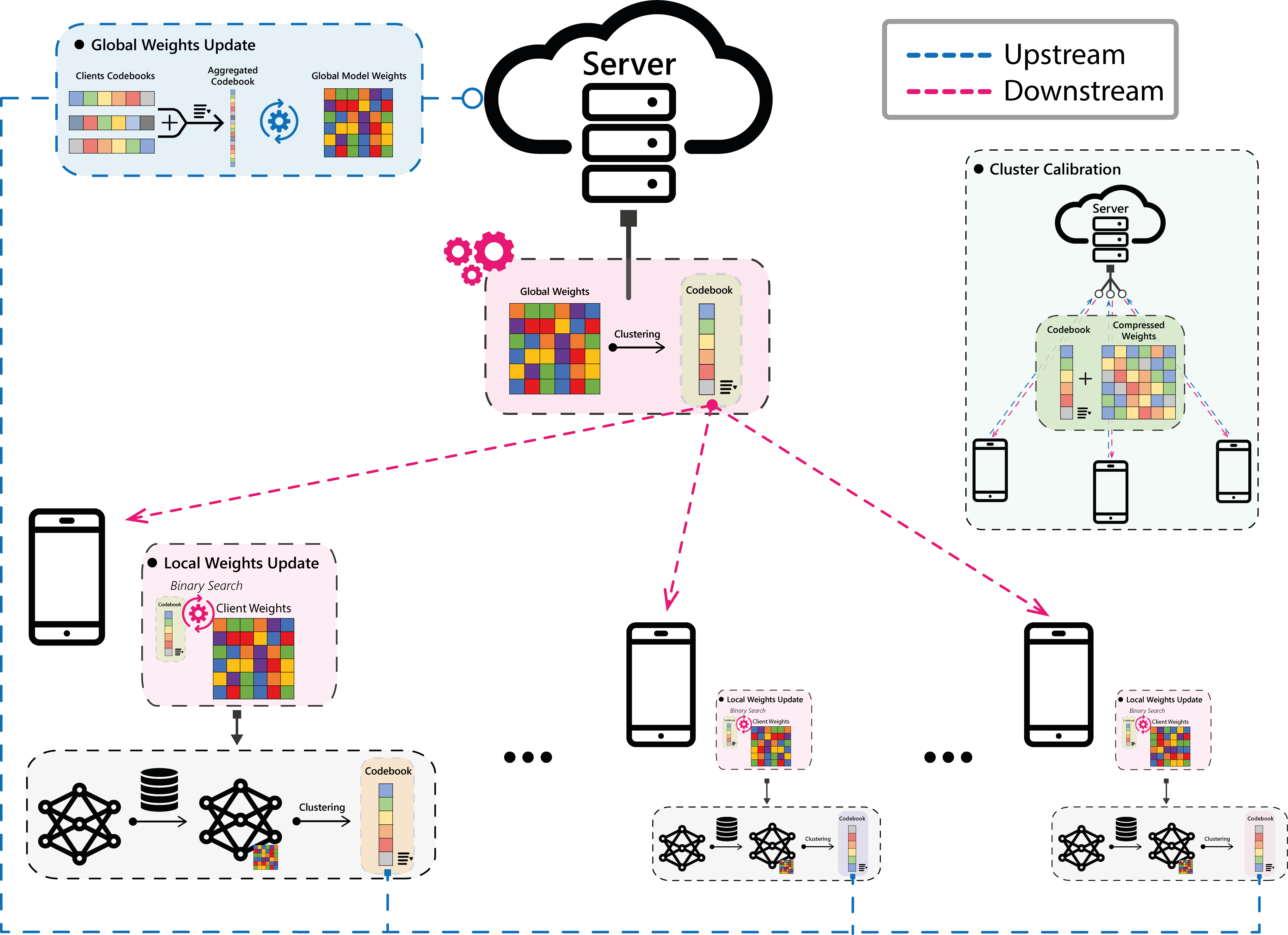}
    \vspace{1em}
    \caption{\small{Illustration of communication-efficient FL via transferring Codebooks.~\method~comprises two stages: 1) Downstream data transmission reduction (in red); Server computes the codebook and transmits it to clients. Clients update their model weights using a computationally inexpensive binary search. 2) Upstream data transmission reduction (in blue); Clients extract their codebook from their locally trained model, while the server updates global model weights based on the aggregated codebook. To calibrate clusters, periodically the compressed weights are transmitted together with the codebook.} \label{fig:overview}}
    \vspace{1em}
\end{figure*}

While previous weight clustering approaches effectively compress the exchanged messages, the transmission of both codebooks and compressed weight matrices at each federated round restricts their potential for substantial reduction in data volume. Furthermore, most existing approaches focus on limiting the upstream communication route~\cite{ fedpaq, fedsparsify,  fedzip, top_k}, and often impose significant computational overhead on the clients. For example, gTop-k~\cite{gtop_k} requires sorting of the weight matrices, while MUCSC~\cite{mucsc} and FedZip~\cite{fedzip} applies clustering in each layer of the model. In response to the aforementioned concerns, ~\method~primarily transfers codebooks during the FL training process, achieving significant reduction in utilized bandwidth and volume of exchanged messages in the bidirectional communication route, while imposing minimal computational overhead for clients. Specifically,~\method~involves a single K-means clustering algorithm execution for clients' weight matrices and a computationally inexpensive binary search for weight reconstruction from the received codebook. Table~\ref{tab:related_work} presents a concise comparison between~\method~and other state-of-the-art communication-efficient FL schemes.

\section{Methodology}
\input{methodology}

\section{Experiments}

In this section,  we outline the evaluation setup, including all considered federated parameters, baselines, and metrics used to evaluate and compare against our method.

\subsection{Experiment Setup\label{ssec:setup}}

\noindent \textbf{Models and Dataset.} 
For performance evaluation, we use publicly available datasets of a wide range of classification tasks from both vision and audio domains. To ensure comparability, we use standard training/test splits, as provided by the original datasets, for all datasets. From the vision domain, we use the CIFAR-10~\cite{krizhevsky2009learning} and CIFAR-100~\cite{krizhevsky2009learning}, where the tasks of interests are object detection among 10 and 100 classes, respectively. As backbone model architectures for those tasks, we use two backbone architectures, namely ResNet-20~\cite{he2016deep}, MobileNet (v2)~\cite{sandler2018mobilenetv2}, covering two of the most widely explored deep learning model architectures in on-device learning thanks to their relatively compact model size. From the audio domain, we use SpeechCommands (v2) dataset~\cite{spcm}, where the learning objective is to detect when a particular keyword is spoken out of a set of twelve target classes, and utilize YAMNet~\cite{yamnet} (an adaptation of MobileNet for audio modality), which is suitable for edge devices~\cite{yamnet_edge}. Lastly, in all our experiments, we utilize Adam optimizer with a learning rate of $0.001$.

\begin{table}[!t]
    \scriptsize \centering
    \captionsetup[table]{justification=centering}
    \caption{\small{Details from our experimental setup.}\label{tab:setup}}
        \begin{tabular}{@{}lcc@{}}
            \toprule
            \multicolumn{1}{c}{\textbf{Name}} & \textbf{Parameter} & \textbf{Range} \\ 
            \midrule[0.5pt] 
            Number of Clients                               & $N$       & 10 \\
            Number of Federated Rounds                      & $R$       & 60 \textemdash 1000 \\
            Minimum Round for Transferring Codebook         & $R_{cb}$  & 2 \textemdash 4 \\
            Number of Local Train Steps                     & $E$       & 2 \textemdash 8 \\
            Number of Clusters                              & $K$       & 16 \textemdash 128 \\
            Clients Participation Rate                      & $\rho$    & 10 \textemdash 100\% \\
            Client class concentration Rate                 & $C_{p}$   & 10 \textemdash 100\% \\ 
            Compressed Weights Upstream Transfer Rate       & $F_1$     & 0 \textemdash 100\% \\
            Compressed Weights Downstream Transfer Rate     & $F_2$     & 0 \textemdash 100\% \\
            Dirichlet distribution Concentration            & $\beta$   & 0.1 \textemdash 10 \\
            \bottomrule
        \end{tabular}
\end{table}

\noindent \textbf{Federated Settings.} We control the federated settings with the following parameters: \begin{inparaenum}[1)] \item  number of clients - $N$, \item client train epochs per round - $E$, and \item number of rounds - $R$, \item clients' participation rate per round - $\rho$, \item client class concentration - $C_{p}$\end{inparaenum}. In~\method, we introduce four hyperparameters to control the trade-off between accuracy and data transmission reduction: \begin{inparaenum}[1)] \item number of clusters - $K$, and \item minimum round for codebook transfer - $R_{cb}$ - \item compressed weights upstream transfer rate - $F_1$, \item compressed weights downstream transfer rate - $F_2$ \end{inparaenum}. The ``\textit{wordlength}'' for measuring the data transmission reduction is set to 32, since PyTorch’s floating-point values have a word length of 32.  To ensure the stability of the training procedure in FL (e.g., clusters are not selected based on random weights), we initiate the transferring of codebooks after the second and fourth federated rounds for the ResNet-20 and the MobileNet models, respectively ($R_{cb}$=$2$/$4$). An overview of our evaluation setup parameters is shown in Table~\ref{tab:setup}.
For the data distribution process in our federated experiments, we explored both IID ($C_{p} \approx 1$) and non-IID ($C_{p}<<1$) distributions, where we partitioned our datasets across the available clients in a non-overlapping fashion, controlling both the amount of data across clients and the classes availability across clients. Specifically, for IID settings we use a Dirichlet distribution with concentration parameter $\beta$=$10.0$ (noted as $Dir(10.0)$) over classes, resulting in $C_{p} \approx 1$, while in non-IID setting we use a $\beta$ concentration of $0.1$ that produces a $C_p \approx 0.1$ (meaning that 10\% of total classes were available per client). It is worth mentioning that even if the meaning of non-IID is generally straightforward, data can be non-IID in many ways. For a concise review of the various non-IID distribution schemes in FL, we refer an interested reader to~\cite{OpenChallenges}. In our work, the term non-IID describes a data distribution with both a label distribution skew and a quantity skew. These types of data distributions are common across clients' data in federated settings~\cite{OpenChallenges}, where each client frequently corresponds to a particular user (affecting the label distribution), and the application usage across clients can differ substantially (affecting the amount of data among clients). For a fair comparison between our experiments, we managed any randomness during data partitioning and training procedures with a seed value. 

\noindent \textbf{Evaluation Metrics and Baselines.} 
We performed experiments utilizing standard FL (\textit{FedAvg}~\cite{fedavg}) to compared our considered baselines and~\method~in terms of data transmission reduction and achieved accuracy, similar to~\cite{fedzip}. From the related weight clustering approaches, it is worth noticing that we were unable to reproduce MUCSC~\cite{mucsc} since no details of the underlying clustering algorithm was reported. Additionally, we performed experiments where we combined weight clustering with \textit{FedAvg} (noted as \textit{FedAvg$_{ws}$}), where both client and global model updates were compressed prior to transmission. Furthermore, we performed experiments using FedZip~\cite{fedzip} where we set $K$=$15$, which we find to work well for the considered architectures and tasks in our preliminary experiments. In all our experiments, we considered Top-1 classification accuracy on test set, while the accuracy loss ($\delta$-Acc) from standard FL (\textit{FedAvg}) is reported for each baseline for ease of comparison. Furthermore, for~\method~we report the DTR ratio based on Equation~\ref{eq:dtr}. 

\begin{table*}[!htb]
    \small \centering
    \caption{\small{Performance evaluation of~\method~ in both IID - $C_p\approx1.0$ using $Dir(10.0)$ over classes - and non-IID - $C_p \approx 0.1$ using $Dir(0.1)$ over classes - settings across different datasets using (a) ResNet-20, and (b) MobileNet architectures. We report the accuracy loss ($\delta$-Acc) on test set and Data Transmission Reduction (DTR) ratio for downstream, upstream and joint communication routes compared with \textit{FedAvg}. Detailed federated parameters are presented in Tables~\ref{tab:resnet20_hp}-\ref{tab:mobilenet_hp} of Appendix~\ref{sec:appendix_c}.}\label{tab:main}}
    \begin{subtable}[t]{0.49\textwidth}
        \resizebox{1.0\textwidth}{!}{%
            \begin{tabular}{@{}clcccccccc@{}}
                \toprule
                    \multicolumn{1}{l}{\multirow{2}{*}{}} 
                    & \multirow{2}{*}{\textbf{Dataset}} 
                        & \multicolumn{2}{c}{\textbf{\textit{IID} ($C_p\approx1.0$)}} 
                        & \multicolumn{2}{c}{\textbf{\textit{non-IID} ($C_p\approx0.1$)}} 
                        & \multirow{2}{*}{\textit{\textbf{Down$_{DTR}$}}}
                        & \multirow{2}{*}{\textit{\textbf{Up$_{DTR}$}}}
                        & \multirow{2}{*}{\textit{\textbf{Total$_{DTR}$}}} 
                    \\ \cmidrule(l{.01\linewidth} r{.01\linewidth}){3-4} \cmidrule(l{.01\linewidth} r{.001\linewidth}){5-6}
                    &  & \textit{\textbf{Accuracy}} & \textit{\textbf{$\delta$-Acc}}  & \textit{\textbf{Accuracy}} & \textit{\textbf{$\delta$-Acc}} & & & \\ 
                \midrule
                    \multirow{2}{*}{$\rho$=$0.1$}
                        & CIFAR-10  & 80.33 & -1.51 & 71.09 & -1.67 & 26 & 10.6 & 15 \\
                        & CIFAR-100 & 54.76 & -3.04 & 48.92 & -3.74 & 26.2 & 10.6 & 15.1 \\ 
                    \midrule
                    \multirow{2}{*}{$\rho$=$1.0$}
                        & CIFAR-10  & 88.67 & -1.43 & 85.03 & -1.07 & 22.8 & 10.3 & 14.2 \\ 
                        & CIFAR-100 & 61.22 & -3.15 & 54.33 & -3.19 & 24.2 & 10.4 & 14.6 \\
                \bottomrule
                \newline \newline
            \end{tabular}%
        }
        \caption{ResNet-20 architecture\label{tab:main_resnet}}
    \end{subtable}
    \hfill
    \begin{subtable}[t]{0.49\textwidth}
        \resizebox{1.0\textwidth}{!}{%
            \begin{tabular}{@{}clcccccccc@{}}
                \toprule
                    \multicolumn{1}{l}{\multirow{2}{*}{}} 
                    & \multirow{2}{*}{\textbf{Dataset}} 
                        & \multicolumn{2}{c}{\textbf{\textit{IID} ($C_p\approx1.0$)}} 
                        & \multicolumn{2}{c}{\textbf{\textit{non-IID} ($C_p\approx0.1$)}} 
                        & \multirow{2}{*}{\textit{\textbf{Down$_{DTR}$}}}
                        & \multirow{2}{*}{\textit{\textbf{Up$_{DTR}$}}}
                        & \multirow{2}{*}{\textit{\textbf{Total$_{DTR}$}}} 
                    \\ \cmidrule(l{.01\linewidth} r{.01\linewidth}){3-4} \cmidrule(l{.01\linewidth} r{.001\linewidth}){5-6}
                    &  & \textit{\textbf{Accuracy}} & \textit{\textbf{$\delta$-Acc}}  & \textit{\textbf{Accuracy}} & \textit{\textbf{$\delta$-Acc}} & & & \\ 
                \midrule
                    \multirow{2}{*}{$\rho$=$0.1$}
                        & CIFAR-10  & 86.38 & -0.33 & 78.79 & -0.81 & 13.7 & 9.1 & 10.9 \\
                        & CIFAR-100 & 67.83 & -1.93 & 54.97 & -2.91 & 13.7 & 9.1 & 10.9 \\ 
                    \midrule
                    \multirow{2}{*}{$\rho$=$1.0$}
                        & CIFAR-10  & 92.03 & -0.04 & 89.14 & -0.28 & 12.6 & 8.7 & 10.3 \\ 
                        & CIFAR-100 & 70.36 & -1.67 & 62.01 & -2.67 & 13.2 & 8.9 & 10.6 \\
                \bottomrule
                \newline \newline
            \end{tabular}%
        }
        \caption{MobileNet architecture\label{tab:main_mobilenet}}
    \end{subtable}
\end{table*}

\begin{table}[!h]
    \small \centering
    \caption{\small{Performance evaluation of~\method~on SpeechCommands~\cite{spcm} using YAMNet~\cite{yamnet} architecture in IID setting - $C_p \approx 1.0$ using $Dir(10.0)$ over classes . We report \textit{FedAvg} accuracy, the accuracy loss ($\delta$-Acc) compared to \textit{FedAvg} on test set, and Data Transmission Reduction (DTR) ratio for downstream, upstream, and joint communication routes compared with \textit{FedAvg}. Detailed federated parameters are presented in Table~\ref{tab:yamnet_hp} of Appendix~\ref{sec:appendix_c}.\label{tab:spcm_yamnet}}}
    \resizebox{0.85\columnwidth}{!}{%
        \begin{tabular}{ccccccc}
            \toprule
            & \textit{\textbf{Accuracy}} & \textit{\textbf{$\delta$-Acc}} & \textit{\textbf{Down$_{DTR}$}} & \textit{\textbf{Up$_{DTR}$}} & \textit{\textbf{Total$_{DTR}$}} \\
            \midrule
            $\rho$=$0.1$ & 91.02 & -0.36 & 13.7  &  9.1  & 10.9  \\
            $\rho$=$1.0$ & 95.75 & -0.17 & 12.8 & 8.7 & 10.3 \\
            \bottomrule
        \end{tabular}%
    }
\end{table}

\begin{figure*}
    \centering \small
    \captionsetup[subfigure]{justification=centering}
    \begin{subfigure}[b]{0.45\textwidth}
        \centering
        \includegraphics[width=\linewidth]{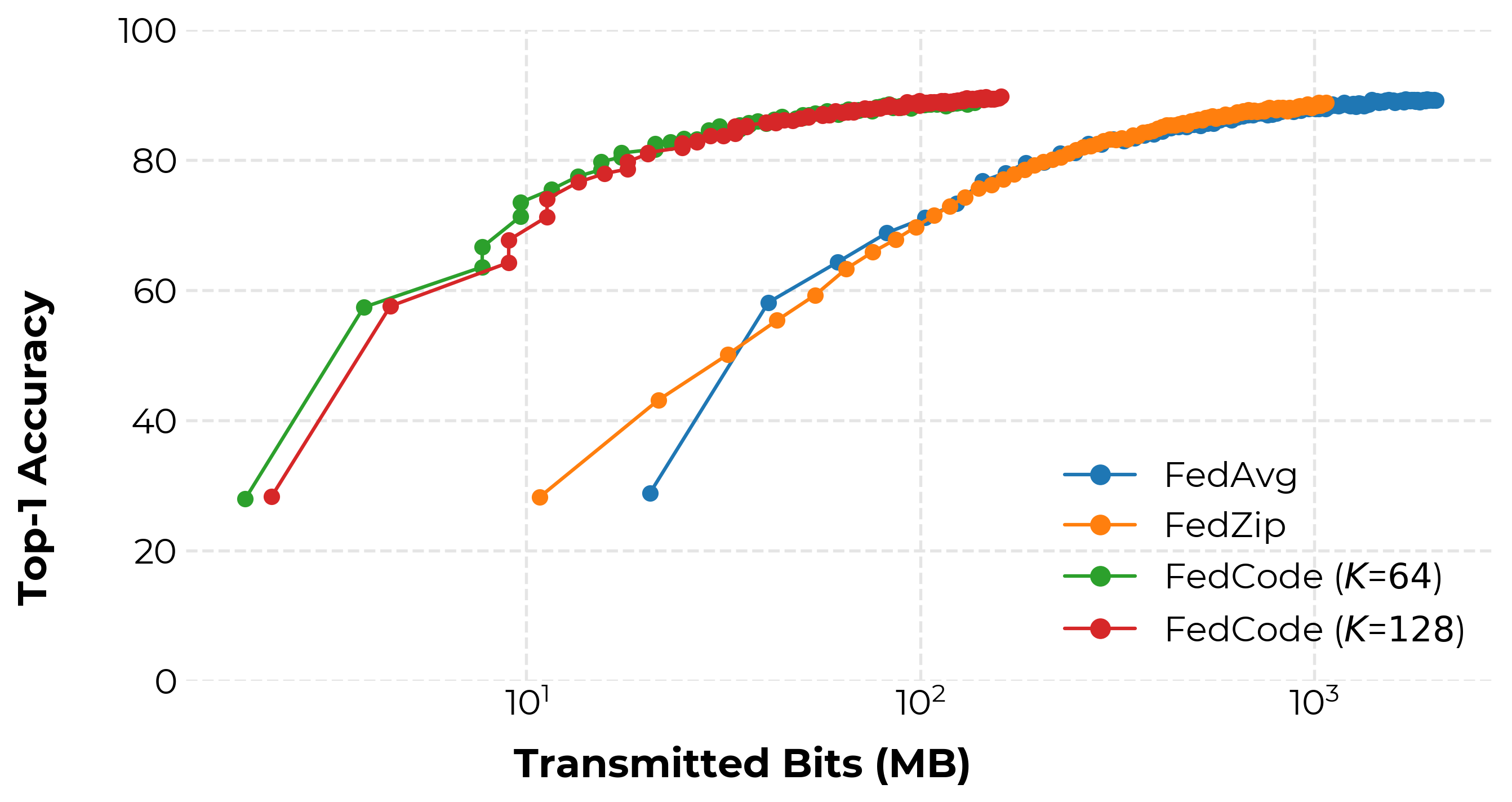}
        \vspace{0.1cm}
        \caption{\small{CIFAR-10 on IID settings}\label{fig:dtr_acc_iid}}
    \end{subfigure}
    \hspace{0.5cm}
    \begin{subfigure}[b]{0.45\textwidth}
        \centering
        \includegraphics[width=\linewidth]{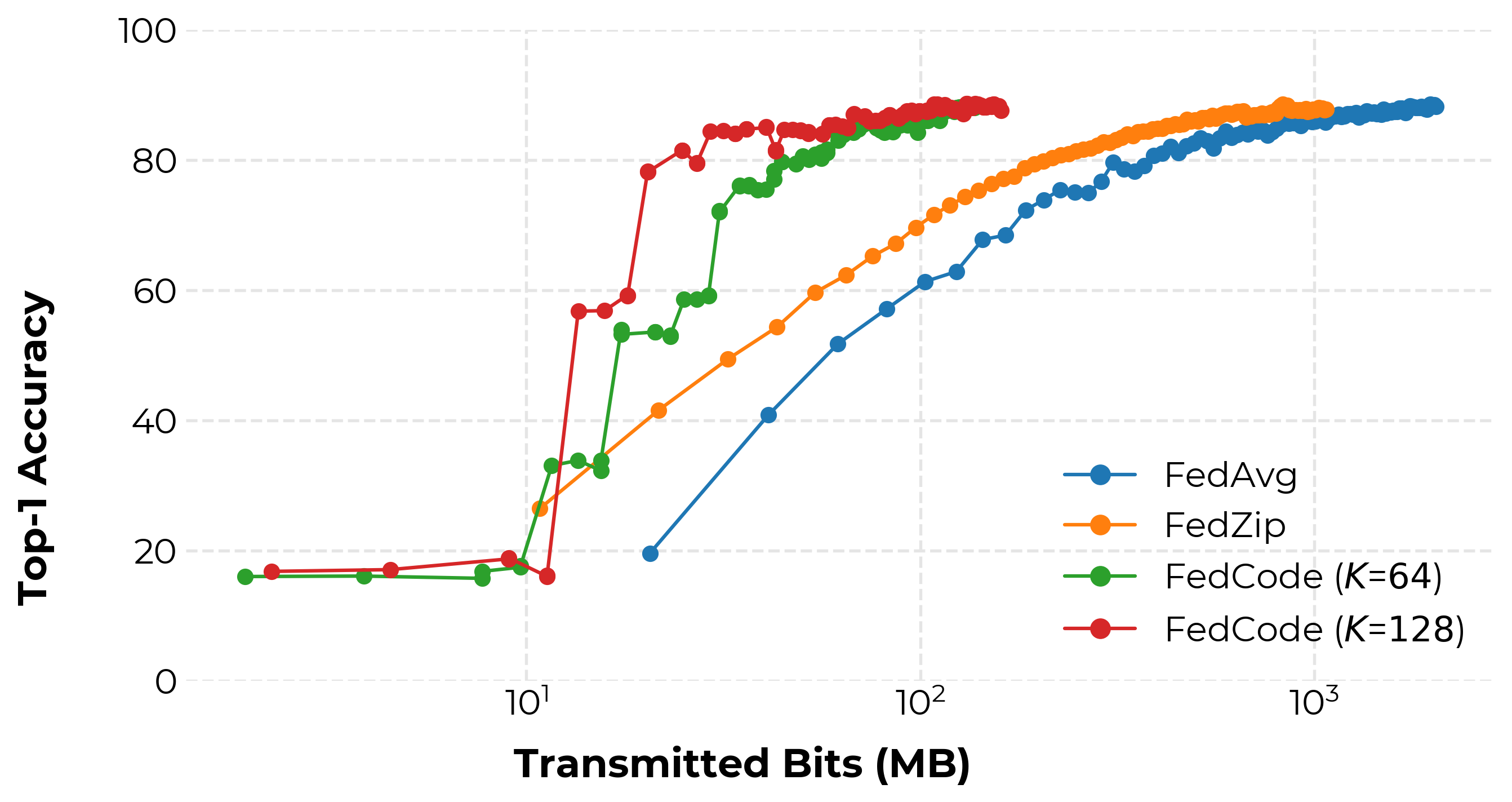}
        \vspace{0.1cm}
        \caption{\small{CIFAR-10 on non-IID settings}\label{fig:dtr_acc_noniid}}
    \end{subfigure}
    \caption{\small{Accuracy versus Volume of Communicated Data trade-off for~\method~using ResNet-20 architecture in CIFAR-10 for (a) IID, and (b) non-IID settings. In both figures, the transmitted bits are reported in logarithmic scale for ease of comparison. We present the transmitted bits in linear scale in Figure~\ref{fig:dtr_acc_linear} of Appendix~\ref{sec:appendix_d}.}\label{fig:dtr_acc}}
\end{figure*}

\section{Results}
In this section, we present performance evaluation of FedCode across a wide range of data distributions and tasks, as described in Section~\ref{ssec:setup}. Our goal is to measure the effectiveness of~\method~in training federated models in terms of communication costs and bandwidth utilization. Moreover, as efficient utilization of the available bandwidth is critical in edge devices, we provide an analysis of bitrate reduction of~\method~in Appendix~\ref{sec:appendix_d}.

\subsubsection*{\textbf{Performance in IID data distributions}}
To assess the effectiveness of~\method~in reducing the communication cost of FL, we first performed experiments under IID data distribution across all considered datasets and model architectures. We measured the DTR and model accuracy compared to standard FL. The results are presented in Tables \ref{tab:main_resnet} and \ref{tab:main_mobilenet}, from which one may notice that~\method~effectively reduces communication costs while maintaining a high global model accuracy. Specifically, using ResNet-20 on CIFAR-10,~\method~achieved an average DTR of 14.2 and 15 with $\rho$=$1$ and $\rho$=$0.1$, while maintaining a highly accurate model with accuracy reductions ($\delta$-Acc) of $-1.43\%$ and $-1.51\%$, respectively. Using ResNet-20 on CIFAR100, the results were similarly impressive, with \method~achieving an average DTR of 14.6 and 15.1 for $\rho$=$1$ and $\rho$=$0.1$, while maintaining a comparable level of accuracy, as indicated by a $\delta$-Acc of $-3.19\%$ and $-3.04\%$, respectively. In our experiments with MobileNet, which is a larger model compared to ResNet-20,~\method~achieved a DTR of $10.3$ and $10.9$ when $\rho$=$1$ and $\rho$=$0.1$, respectively, while also maintaining high accuracy with $\delta$-Acc of $-0.04\%$ and $-0.33\%$, respectively. Similarly, when applying~\method~to MobileNet on CIFAR-100, we obtained a DTR of $10.6$ and $10.9$ for $\rho$=$0$ and $\rho$=$0.1$, respectively, while keeping the model's performance on par with the standard FL models (i.e., achieving $\delta$-Acc of $-1.67\%$ and $-1.93\%$). 

In addition, we assessed the~\method~'s performance under audio classification tasks using YAMNet~\cite{yamnet} architecture on  SpeechCommands~\cite{spcm} dataset, as detailed in Table \ref{tab:spcm_yamnet}. It is worth noting that a similar pattern like MobileNet and ResNet-20, emerged, where~\method~achieved a DTR of $10$ with minimal impact on the accuracy of the models. From this, we conclude that~\method~can effectively reduce the amount of transmitted data during FL training while maintaining a similar/comparable performance to the standard FL process, which holds true across different types of classification tasks.

\subsubsection*{\textbf{Performance in non-IID data distributions}}
To assess the impact of different data distributions on~\method~performance, we conducted experiments using ResNet-20 and MobileNet on CIFAR-10 and CIFAR-100 under non-IID data distribution. Table \ref{tab:main} represents the highest global model accuracy achieved in settings with non-IID data distribution. When we applied ResNet-20 to CIFAR-10, the~\method~demonstrated remarkable accuracy, resulting in an accuracy reduction of $-1.07\%$ and $-1.67\%$ for $\rho$ values of $1$ and $0.1$, respectively. Similarly, for ResNet-20 on CIFAR-100,\method~achieved a comparable level of accuracy with accuracy reductions of $-3.19\%$ and $-3.74\%$ for $\rho$ values of $1$ and $0.1$. Our experiments with MobileNet showed that\method~achieved high accuracy with accuracy losses of $-0.28\%$ and $-0.81\%$ for $\rho$ values of $1$ and $0.1$, respectively. Furthermore, when\method~was applied to MobileNet on CIFAR-100, the model's performance resulted in an accuracy loss of $-2.67\%$ and $-2.91\%$ for $\rho$ values of $0$ and $0.1$, respectively. As shown in Table \ref{tab:main}, the corresponding data transmission reduction indicates the effectiveness of the proposed approach in reducing data transmission while handling different data distributions, where it can achieve competitive accuracy.

\subsubsection*{\textbf{Impact of Codebook Size on Accuracy}}

\begin{figure}[t]
  \centering
  \includegraphics[width=0.45\textwidth]{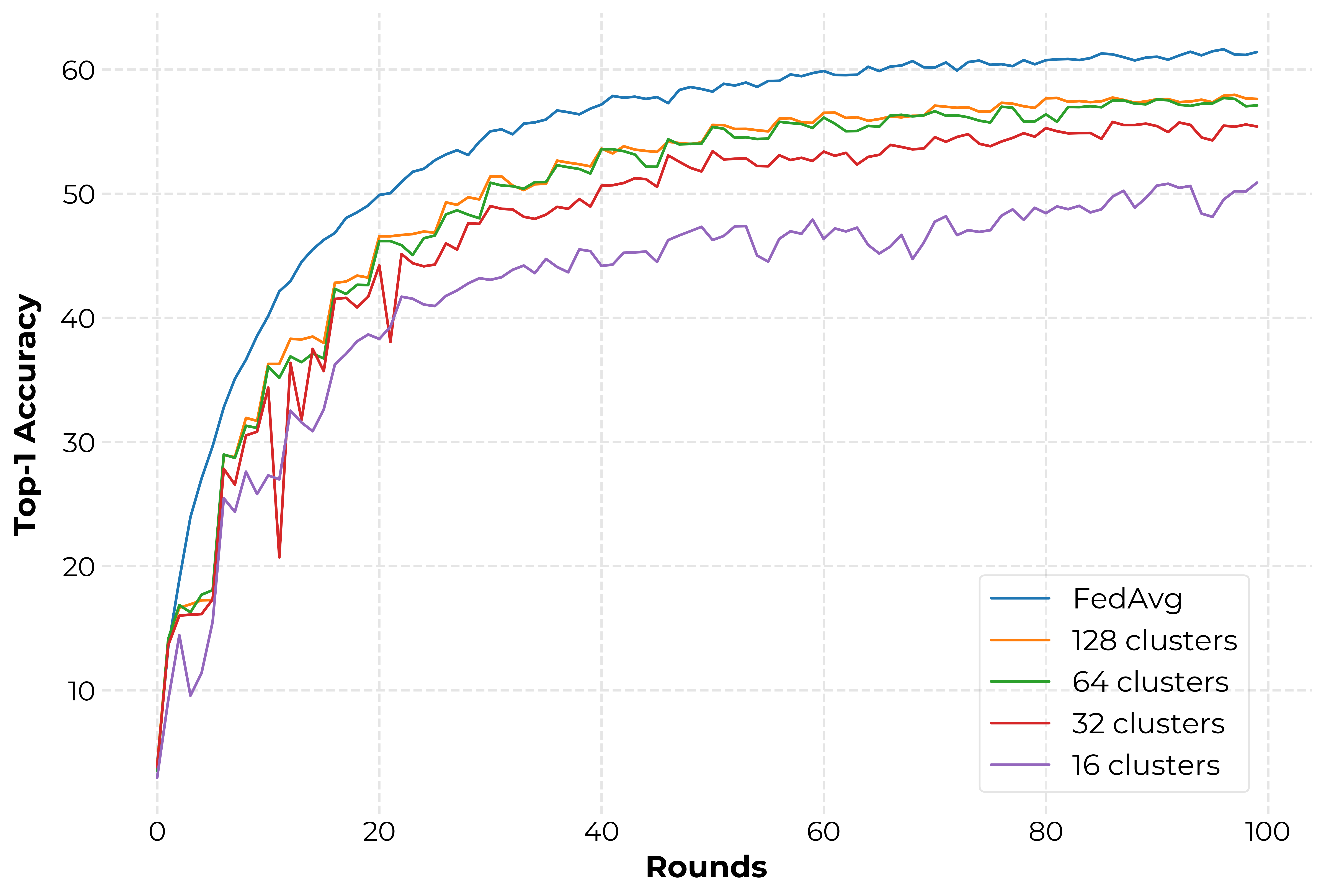}
  \caption{\small{Effect of the number of clusters on model accuracy for Resnet-20 on CIFAR-100. The cluster numbers considered are $16$, $32$, $64$, and $128$. For~\method, the hyperparameters are set to $F_1$=$0.2$, $F_2$=$0.5$, and $R_{cb}$=$2$. Remaining federated parameters are set to $R$=$100$, $C$=$10$, $P_{r}$=$100\%$, and $E$=$4$.\label{fig:cluster_accuracy}}}
\end{figure}

To investigate the impact of the communicated number of clusters ($K$) on global models' accuracy, we performed experiments varying $K$ from 16 to 128. Figure~\ref{fig:cluster_accuracy} presents our finding, which we indicates increasing $K$ leads to increase of accuracy. However, this increase quickly saturates once number of clusters increases above $64$ clusters. Hence, increasing $K$ to extremely high values has no meaningful effect on the models' performance. Furthermore, revisiting Equation~\ref{eqn:dtr}, one may notice that increasing $K$ has a negligible effect on DTR due to the logarithmic relationship between the cluster number and the data transmission reduction, as demonstrated in Section~\ref{sc:dtr_anlysis}.

\subsubsection*{\textbf{Impact of Local Training Steps  on Accuracy}}

\begin{figure}[t]
  \centering
  \includegraphics[width=0.45\textwidth]{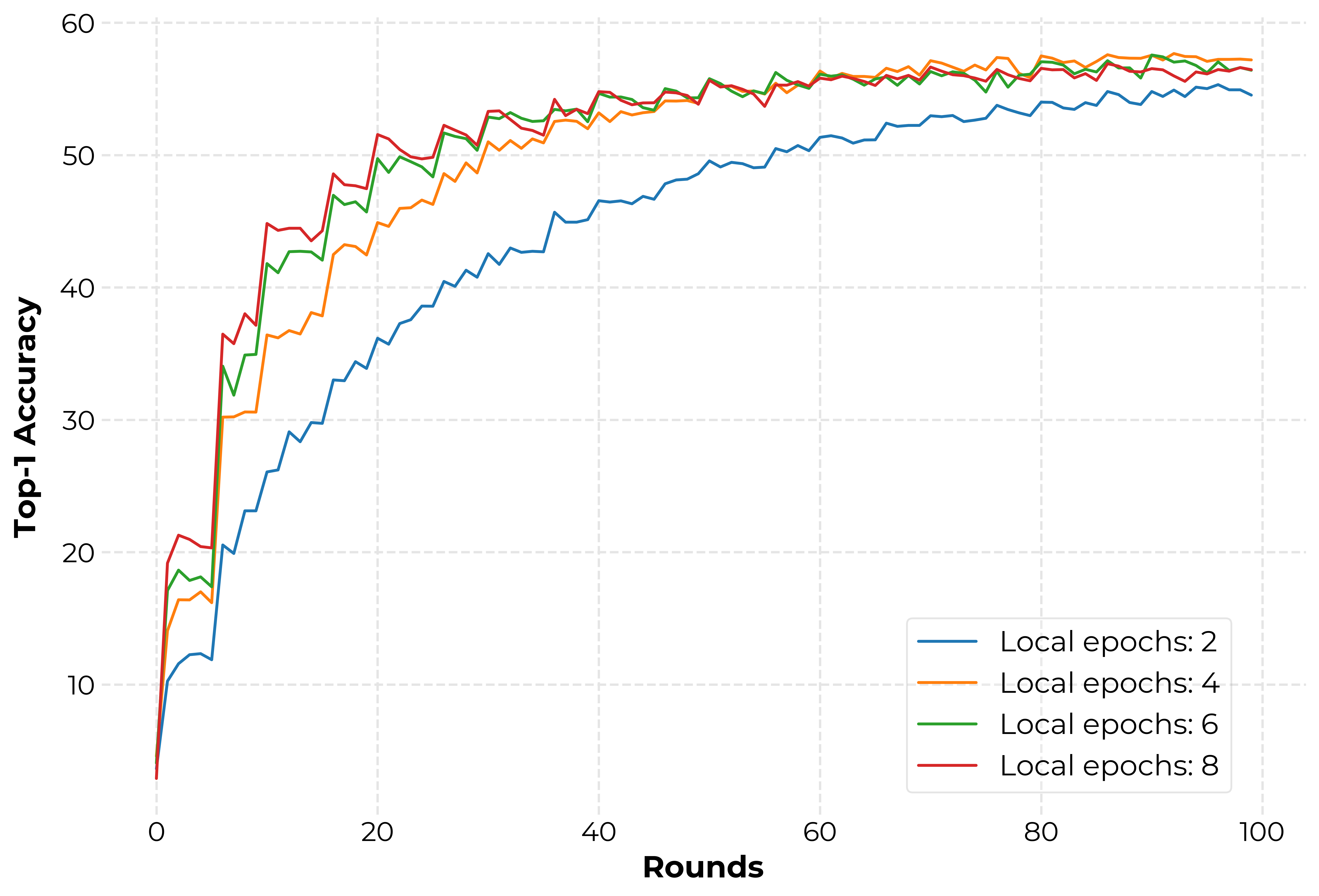}
  \caption{\small{Effect of the number of local epochs on model accuracy for Resnet-20 on CIFAR-100. The local epochs considered are $2$, $4$, $6$, and $8$. For~\method, the hyperparameters are set to $F_1$=$0.2$, $F_2$=$0.5$, and $R_{cb}$=$2$. Remaining federated parameters are set to $R$=$100$, $C$=$10$, $P_{r}$=$100\%$, and $E$=$4$.\label{fig:local_epochs_accuracy}}}
\end{figure}

To explore the impact of the number of local training steps (i.e., local epochs, referred to as $E$) on the accuracy of global models, we conducted experiments by altering the value of $E$ ranging from 2 to 8. The results, as depicted in Figure~\ref{fig:local_epochs_accuracy}, reveal that increasing the value of $E$ leads to an improvement in accuracy. However, although increasing  $E$ beyond four epochs initially boosted accuracy, it did not result in any additional enhancements of the final accuracy. Therefore, increasing $E$ to high values has no meaningful effect on the models' performance.

\subsubsection*{\textbf{Exploring Hyperparameters Space}}

\begin{table}[b]
    \small \centering
    \caption{\small{~\method~evaluation across different ranges of $F_1$ and $F_2$. Data Transmission Reduction (DTR) ratio and $\delta$-Acc (accuracy degradation) compared to \textit{FedAvg}, along with the amount of transferred bits, measured in MB for ResNet-20 on CIFAR-100 is reported. For~\method, we use $K$=$64$ and $R_{cb}$=2. Remaining federated parameters are set to $R$=$100$, $C$=$10$, $P_{r}$=$100\%$, and $E$=$4$. \textit{FedAvg} transmitted bits totaled $2102.44$$MB$.\label{tab:f1_2_dtr}}}
    \begin{tabular}{lcccc}
        \toprule
        \textbf{F1} & \textbf{F2} & \textbf{$\delta$-Acc} & \textbf{DTR}& \textbf{Transmitted Bits (MB)}\\
        \midrule
        1 & 1 & -0.86 & 5.3 &  394.21 \\
        0.5 & 0.5 & -3.46 & 10.4 & 201.08 \\
        0.2 & 0.5 & -3.15 & 14.6 &  143.92\\
        0.33 & 0.33& -5.94 & 15.2 & 138.01   \\
        0.1 & 0.1 & -21.64 & 44.3 & 47.35 \\
        0 & 0 & -46.88 & 264.7 & 7.9 \\
        \bottomrule
    \end{tabular}
\end{table}

\begin{figure}[t]
  \centering
  \includegraphics[width=0.45\textwidth]{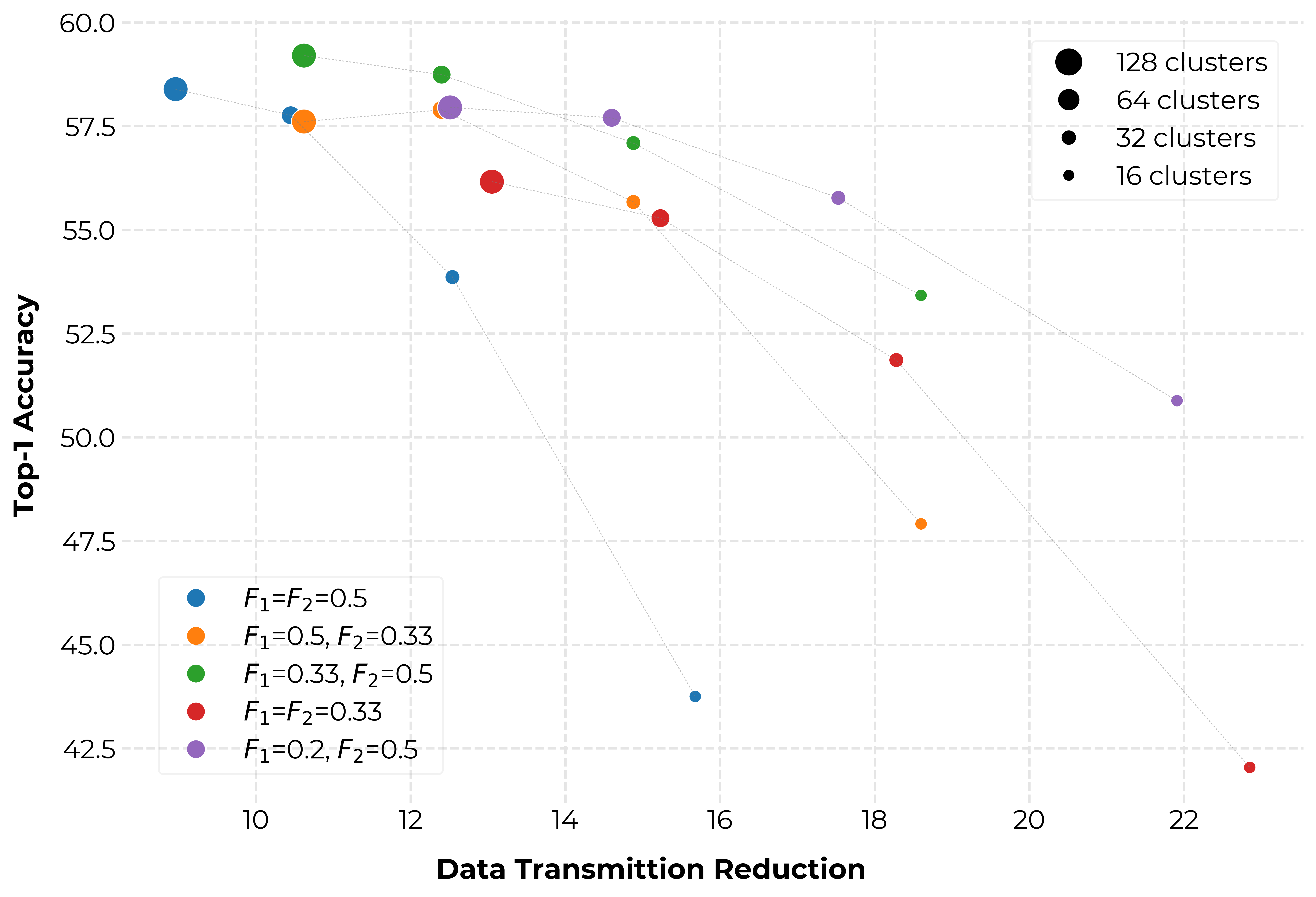}
  \caption{\small{Trade-off between model accuracy, number of clusters ($K$), and the frequency of transferring compressed weights in downstream and upstream communication ($F_1$, $F_2$), for $20$ different settings with Resnet-20 on CIFAR-100.  Data Transmission Reduction (DTR) compared to \textit{FedAvg} and model accuracy for each setting is reported. The range of $K$ was varied from $16$ to $128$, and the values of $F_1$ and $F_2$ were varied from $0.2$ up to $0.5$. Remaining federated parameters are set to $R$=$100$, $C$=$10$, $P_{r}$=$100\%$, and $E$=$4$.
 \label{fig:accuracy__cluster_r_ccr}}}
\end{figure}

To assess the trade-off between DTR and global model accuracy, we conducted experiments for a range of $K$ from 16 to 128, where we varied the frequency of cluster calibration in downstream and upstream communications. Figure~\ref{fig:accuracy__cluster_r_ccr} illustrates the resulting model accuracy in combination with DTR for the considered hyperparameters space. One may notice that while increasing both $K$ and $F_{1/2}$ has a beneficial effect on model accuracy, the best model performance is found for $F_1$=$0.33$ and $F_2$=$0.5$, even surpassing the accuracy obtained when $F_1$=$F_2$=$0.5$. This observation indicates that transferring only the codebook from both the server and clients in the same round can degrade the global model's accuracy. In other words, the best model performance is achieved when the transmission of compressed weights for upstream and downstream do not occur simultaneously (i.e., when the server transmits the compressed weights, clients only transfer codebooks).

Furthermore, considering that the transmission rates of compressed weights have a substantial influence on the accuracy and DTR, as depicted in Table~\ref{tab:f1_2_dtr},  we investigated accuracy, DTR, and the amount of transferred bits across various settings within the F1 and F2 range, which varied from 0 to 1. The results demonstrated that decreasing the frequency of compressed weight transfers led to a substantial rise in DTR. However, this decrease also resulted in a reduction of the global model's accuracy.

\subsubsection*{\textbf{Comparison with Baselines}}

\begin{table}[b]
    \centering
    \caption{\small{Comparison of~\method~with existing weight clustering approaches using ResNet-20 on CIFAR-100. Downstream, Upstream, and joint Data Transmission Reduction (DTR) ratio compared to \textit{FedAvg} and model accuracy (\%) is reported for each baseline. For~\method, we use $K$=$64$, $F_1$=$0.2$ and, $F_2$=$0.5$. Remaining federated parameters are set to $R$=$100$, $C$=$10$, $P_{r}$=$100\%$, and $E$=$4$.\label{tab:comparison}}}
    \resizebox{0.95\hsize}{!}{%
    \begin{tabular}{lccccc}
        \toprule
        Method & Accuracy & Downstream DTR & Upstream DTR & DTR \\
        \midrule
        FedAvg                  & 61.22 & -    & -      & - \\
        FedAvg$_{ws}$           & 58.78 & 5.3  & 5.3    & 5.3 \\
        FedZip \cite{fedzip}    & 57.98 & -    & 19.8   & 1.9 \\
        FedCode                 & 58.07 & 24.2 & 10.4   & 14.6 \\
        \bottomrule
    \end{tabular}%
    }
\end{table}

Table~\ref{tab:comparison} presents global model performance for each of the considered baselines, namely \textit{FedAvg}, \textit{FedAvg} with weight clustering (referred as \textit{FedAvg$_{ws}$}) and FedZip~\cite{fedzip} for the CIFAR-100 dataset. One may notice that by transferring compressed weights via \textit{FedAvg$_{ws}$}, we can achieve a significant DTR of $5.3$ in both upstream and downstream communication routes, while accuracy drops from 61.22\% (reported in \textit{FedAvg} using the same dataset) to 58.78\% . Additionally, FedZip, which introduces Top-k pruning and layer-wise weight clustering to compress individual layers, achieves an accuracy of 57.98\% with an upstream DTR of $19.8$. However, as FedZip only performs compression in the upstream communication route the overall DTR falls to $1.9$ since the complete weight matrices need to be downloaded by clients at each federated round. On the other hand,~\method~achieves an accuracy of $58.07\%$ and demonstrated a significant DTR for both downstream and upstream communication channels. Specifically, we report a downstream DTR of $24.2$ and an upstream DTR of $10.4$, while the overall DTR is $14.6$, surpassing all considered baselines. In addition, Figure \ref{fig:dtr_acc} illustrates a comparison of the volume of transmitted bits between~\method~, \textit{FedAvg}, and \textit{FedZip}, highlighting that the data transferred in~\method~is notably lower than in the other baseline methods. Therefore,~\method~can be considered as a communication-efficient FL, achieving a competitive accuracy to \textit{FedAvg}, while significantly reducing the amount of transmitted data and the utilized bandwidth during training.

\section{Discussion}
Our study aimed to investigate the effectiveness of transferring codebook in reducing data transmission in both downstream and upstream communication. In order to thoroughly examine its impact, we investigated various aspects including the effect on global model accuracy, data transmission efficiency, and the computational overhead on clients. These aspects were selected as they represent key challenges in the context of FL~\cite{OpenChallenges}. As the primary objective of this work was to investigate the effectiveness of transferring codebooks to increase communication-efficiency in FL, we have not investigated model compression techniques to compress deep learning models prior to communication between clients and the server. Nevertheless, it is important to recognize that there are approaches aimed at enhancing compression ratios through the utilization of techniques, such as pruning \cite{ma2019resnet}, quantization, block weight clustering \cite{bingham2022legonet}, or section-wise weight clustering \cite{khalilian2023escepe}. These techniques may potentially be combined with~\method~to further decrease the size of compressed weight matrices and consequently reduce the volume of transmitted data.

\noindent \textbf{Global Model Performance}. Maintaining the performance of the global model is crucial in FL when aiming at reducing communication costs. To achieve a proper balance between data transmission reduction and model accuracy, we carefully controlled three hyperparameters and conducted numerous experiments on diverse classification tasks and different architectures to assess their performance. Tables \ref{tab:main} and \ref{tab:spcm_yamnet} present the results of our experiments, indicating that when every client actively participates in each round the~\method~approach resulted in an accuracy loss of less than $2\%$ for most cases, with the exception of ResNet-20 on CIFAR-100. However, Figure~\ref{fig:accuracy__cluster_r_ccr} illustrates that by adjusting the hyperparameters, we were able to further reduce the accuracy loss. For instance, by fine-tuning our hyperparameters to the frequency of transferring compressed weights for downstream and upstream communication routes to $0.33$ and $0.5$ together with $128$ cluster centers, the accuracy loss for ResNet-20 on CIFAR-100 decreased to $2.03\%$. This demonstrates that with proper fine-tuning of these hyperparameters, \method achieves both high model performance and significant data transmission reduction. Furthermore,~\method~could retain its performance even under non-IID data distribution scenarios, with Resnet-20 reporting an accuracy loss of $1.43$ and $3.15$ and MobileNet reporting an accuracy loss of $0.28$ and $2.67$ on CIFAR10 and CIFAR100, respectively. The results presented in Table \ref{tab:main} also revealed that when the clients' participation rate was adjusted to 10 percent, FedCode still achieved a comparable level of accuracy with an average accuracy reduction of 2.27 for Resnet and 1.13 for MobileNet on IID data distribution, and 2.7 for Resnet and 1.86 for MobileNet on non-IID data distribution. These findings highlight that transferring codebook is effective in reducing data transmission while accommodating diverse data distribution. Lastly, comparison with existing approaches, showcases that~\method~can train models with competitive performance, while providing a superior reduction in data transmission.

\noindent \textbf{Data Transmission Reduction}. Our experimental results demonstrate that~\method~achieved an average 17.8-fold reduction in transmitted data from clients to the server and an average 11.2-fold reduction in transmitted data from the server to clients. Overall, there is an average 12.2-fold reduction in data transmission across the CIFAR-10, CIFAR-100, and SpeechCommands datasets using ResNet-20, MobileNet, and YAMNet architectures, as shown in Tables \ref{tab:main} and \ref{tab:spcm_yamnet}, while maintaining minimal accuracy loss. Our study reveals that it is unnecessary to transfer complete weight matrices after each federated round. For example, in the case of ResNet-20 on CIFAR100, we set the frequency of transferring compressed weights to clients to 0.25, indicating that only the codebook is transferred to clients in approximately 75\% of the federated rounds. Similarly, the frequency of transferring compressed weights to the server was set to 0.5, meaning that for 50\% of the federated rounds, only the codebook is transferred from clients to the server. Additionally, when comparing ~\method~with other weight clustering techniques, it is evident that transferring the codebook results in a superior reduction in data transmission compared to using weight clustering alone, and it also outperforms FedZip, which employs layer-wise weight clustering, as illustrated in Table \ref{tab:comparison}. These findings highlight the potential of the~\method~for significantly reducing communication costs in FL. 

\noindent \textbf{Computational Overhead}. We aimed to minimize the computational overhead on clients, considering their typical resource-constrained nature. Unlike FedZip~\cite{fedzip}, which applied K-means clustering to each layer of the deep learning model individually, we applied K-means clustering once on the entire architecture. This approach reduced the computational burden on clients. During downstream communication, clients only needed to replace their weight values with the nearest cluster center using binary search. Similarly, during upstream communication, clients applied K-means clustering on the entire architecture. Additionally, when the server transferred compressed weights, clients only needed to replace each cluster index in compressed weight matrices with the corresponding center from the codebook. This decompression process had a complexity of $\mathcal{O}\left(P \right)$, resulting in efficient decompression. In contrast, FedZip applied pruning on clients' models, which could impose additional computational requirements on clients. However, by utilizing our approach of one-time clustering, we achieved superior data transmission reduction without imposing significant computational overhead on clients.

\section{Conclusion}

In this study, we proposed~\method, a communication-efficient FL framework that utilizes weight clustering to significantly reduce data transmissions during FL training, while maintaining model performance. Our approach primarily relies on codebook transmission to update clients' models from server and vice versa. Our evaluation across multiple datasets and architectures demonstrates that models trained on~\method~report an average accuracy loss of less than 1.5 compared to standard FL while achieving a $12.2$-fold reduction in data transmission. Experiments under non-IID data distributions, further highlight its ability to handle diverse data distributions and maintain similar performances. Concluding,~\method~showcases promising results in maintaining global model performance and achieving substantial data transmission reductions while minimizing computational overhead for clients, all of which are crucial aspects of FL.

\bibliographystyle{ACM-Reference-Format}
\bibliography{reference}

\clearpage \onecolumn

\appendix

\section{Data Transmission Reduction Bounds}\label{sc:dtr_anlysis}

In this section, we discuss the effectiveness of~\method~in terms of communication-efficiency by analyzing the data transmission reduction ($DTR$) compared to standard FL (\textit{FedAvg}~\cite{fedavg}). Specifically, we measure the ratio between the total number of bits transferred in standard FL across the complete FL training process. Hence, $DTR$ can be defined as follows:

\begin{equation} \label{eq:dtr}
    \resizebox{0.3\hsize}{!}{%
    $
    \frac{ 2 \times P \times \textit{worldlength}}{2 \times K \times \textit{wordlength} + \left ( F_1 + F_2 \right ) \times \left ( P \times  \lceil \log_2 K \rceil \right )}$
    }
\end{equation}

\noindent where $P$ is the number of trainable parameters, $K$ refers to the number of clusters, and $F_1$, $F_2$ are the frequency of downstream and upstream cluster calibration process.

By analyzing the range of $F_1$ and $F_2$, we can identify the upper and lower limits of DTR in~\method~using Equation \ref{eq:dtr}. When $F_1$ and $F_2$ are in the vicinity of zero, we rely solely on codebook transfer at each federated round for the FL process. In this case, the DTR reaches its maximum value of $\frac{P}{K}$. Alternatively, when $F1$ and $F2$ values are close to one, we essentially exploit model compression via weight clustering to reduce the size of exchanged messages. Apart from $F_1$ and $F_2$, the number of clusters ($K$) exhibits a logarithmic effect on the $DTR$, implying that higher values of $K$ lead to marginal improvements in $DTR$.

\begin{figure}[!h]
    \small \centering
    \includegraphics[width=0.4\textwidth]{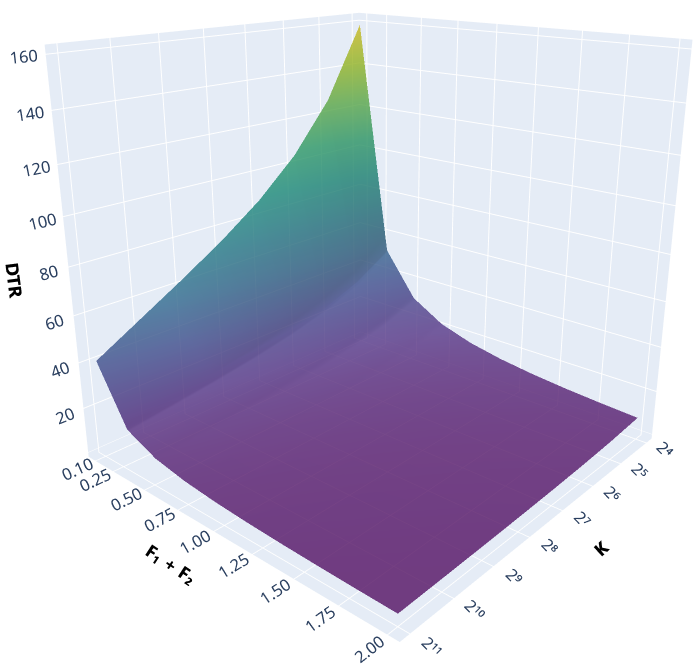}
    \caption{Impact of the frequency of transferring compressed weights ($F_1+F_2$) and the number of clusters ($K$) on the data transmission reduction (DTR).\label{fig:dtr}}
\end{figure}

To clearly illustrate the effect of~\method~hyperparameters in DTR, we consider the joint frequency of cluster calibration occurrences (given by $F_1 + F_2$) in combination with $K$ in Figure~\ref{fig:dtr}. The visualization highlights that the maximum DTR is achieved when both $F_1 + F_2$ and $K$ are low. Conversely, increasing $K$ to a high value does not result in a significant change in the DTR. Therefore, to achieve a high compression ratio, while maintaining global model accuracy, we aim to strike a balance between the number of clusters ($K$) and the frequency of transferring compressed weights ($F_1 + F_2$).

\section{Additional Experimental Details\label{sec:appendix_c}}

In this section, we provide all the detailed experimental details utilized in our evaluation. Clients performed 4 local epochs with a batch size of 128 and a local learning rate of 0.001 using Adam optimizer in all the experiments. We conducted our experiments on NVIDIA Titan X GPUs on an internal cluster server, using 1 GPU per one run. 

For partitioning the datasets across clients, we used the Dirichlet distribution as it is considered an appropriate choice to simulate real-world data distribution across a wide range of researchers~\cite{noniid_1, noniid_2, noniid_3}. Specifically, we sample $p_k \sim Dir_{N} (\beta)$ and allocate a $p_{k,j}$ proportion of the instances of class $k$ to client $j$. Here $Dir(\cdot)$ denotes the Dirichlet distribution and $\beta$ is a concentration parameter ($\beta$>$0$). Using $\beta$=$10$, we simulate a IID federated setup with an approximate class concentration ($C_{p}$) per client close to 1.0 (all classes are available in all clients). Alternatively, with $\beta$=$0.1$, we simulate a highly non-IID FL setting, where clients only hold 10\% of the available classes ($C_{p} \approx 0.1$).

Alternatively, with $\beta$=$0.1$, we simulate a highly non-IID FL setting, where clients only hold 10\% of the available classes ($C_{p} \approx 0.1$). Alternatively, with $\beta$=$0.1$, we simulate a highly non-IID FL setting, where clients only hold 10\% of the available classes ($C_{p} \approx 0.1$).

\begin{table*}[!h]
    \small \centering
    \caption{\small{Hyperparameters used with ResNet20 architecture.}\label{tab:resnet20_hp}}
    \begin{subtable}[t]{0.28\textwidth}
        \resizebox{1.0\textwidth}{!}{%
            \begin{tabular}{ccc}
                \toprule
                \textbf{Parameter} 
                & \begin{tabular}[c]{c}\textbf{\textit{IID}}\\\textbf{($C_p\approx1.0$)}\end{tabular} 
                & \begin{tabular}[c]{c}\textbf{\textit{non-IID}}\\\textbf{($C_p\approx0.1$)}\end{tabular} \\
                \midrule
                \textbf{\textit{params}} & \multicolumn{2}{c}{0.26M} \\
                \textbf{$N$} & \multicolumn{2}{c}{10} \\
                \textbf{$E$} & \multicolumn{2}{c}{4} \\
                \textbf{$\rho$} & \multicolumn{2}{c}{1.0} \\
                \textbf{$lr$} & \multicolumn{2}{c}{1e-3} \\
                \textbf{$R_{cb}$} & \multicolumn{2}{c}{2} \\
                \textbf{$F_1$} & \multicolumn{2}{c}{0.2} \\
                \textbf{$F_2$} & \multicolumn{2}{c}{0.5} \\
                \textbf{$R$} & 60 & 100 \\
                \textbf{$K$} & \multicolumn{2}{c}{64} \\
                \textbf{$Dir(\beta)$} & 10.0 &  0.1 \\
                \bottomrule \newline
            \end{tabular}%
        }
        \caption{$\rho$=$1.0$\label{tab:resnet20_hp_mp}}
    \end{subtable}
    \hspace{1.0cm} %
    \begin{subtable}[t]{0.28\textwidth}
        \resizebox{1.0\textwidth}{!}{%
            \begin{tabular}{ccc}
                \toprule
                \textbf{Parameter} & \begin{tabular}[c]{c}\textbf{\textit{IID}}\\\textbf{($C_p\approx1.0$)}\end{tabular} & \begin{tabular}[c]{c}\textbf{\textit{non-IID}}\\\textbf{($C_p\approx0.1$)}\end{tabular} \\
                \midrule
                \textbf{\textit{params}} & \multicolumn{2}{c}{0.26M} \\
                \textbf{$N$} & \multicolumn{2}{c}{10} \\
                \textbf{$E$} & \multicolumn{2}{c}{4} \\
                \textbf{$\rho$} & \multicolumn{2}{c}{0.1} \\
                \textbf{$lr$} & \multicolumn{2}{c}{1e-3} \\
                \textbf{$R_{cb}$} & \multicolumn{2}{c}{2} \\
                \textbf{$F_1$} & \multicolumn{2}{c}{0.2} \\
                \textbf{$F_2$} & \multicolumn{2}{c}{0.5} \\
                \textbf{$R$} & 600 & 1000 \\
                \textbf{$K$} & \multicolumn{2}{c}{64} \\
                \textbf{$Dir(\beta)$} & 10.0 &  0.1 \\
                \bottomrule \newline
            \end{tabular}%
        }
        \caption{$\rho$=$0.1$.\label{tab:resnet20_hp_lp}}
    \end{subtable}
\end{table*}

\begin{table*}[!h]
    \small \centering
    \caption{\small{Hyperparameters used with MobileNet architecture.}\label{tab:mobilenet_hp}}
    \begin{subtable}[t]{0.28\textwidth}
        \resizebox{1.0\textwidth}{!}{%
            \begin{tabular}{ccc}
                \toprule
                \textbf{Parameter} 
                & \begin{tabular}[c]{c}\textbf{\textit{IID}}\\\textbf{($C_p\approx1.0$)}\end{tabular} 
                & \begin{tabular}[c]{c}\textbf{\textit{non-IID}}\\\textbf{($C_p\approx0.1$)}\end{tabular} \\
                \midrule
                \textbf{\textit{params}} & \multicolumn{2}{c}{2.23M} \\
                \textbf{$N$} & \multicolumn{2}{c}{10} \\
                \textbf{$E$} & \multicolumn{2}{c}{4} \\
                \textbf{$\rho$} & \multicolumn{2}{c}{1.0} \\
                \textbf{$lr$} & \multicolumn{2}{c}{1e-3} \\
                \textbf{$R_{cb}$} & \multicolumn{2}{c}{4} \\
                \textbf{$F_1$} & \multicolumn{2}{c}{0.33} \\
                \textbf{$F_2$} & \multicolumn{2}{c}{0.5} \\
                \textbf{$R$} & 100 & 200 \\
                \textbf{$K$} & 64 & 128 \\
                \textbf{$Dir(\beta)$} & 10.0 &  0.1 \\
                \bottomrule \newline
            \end{tabular}%
        }
        \caption{$\rho$=$1.0$\label{tab:mobilenet_hp_mp}}
    \end{subtable}
    \hspace{1.0cm} %
    \begin{subtable}[t]{0.28\textwidth}
        \resizebox{1.0\textwidth}{!}{%
            \begin{tabular}{ccc}
                \toprule
                \textbf{Parameter} & \begin{tabular}[c]{c}\textbf{\textit{IID}}\\\textbf{($C_p\approx1.0$)}\end{tabular} & \begin{tabular}[c]{c}\textbf{\textit{non-IID}}\\\textbf{($C_p\approx0.1$)}\end{tabular} \\
                \midrule
                \textbf{\textit{params}} & \multicolumn{2}{c}{2.23M} \\
                \textbf{$N$} & \multicolumn{2}{c}{10} \\
                \textbf{$E$} & \multicolumn{2}{c}{4} \\
                \textbf{$\rho$} & \multicolumn{2}{c}{0.1} \\
                \textbf{$lr$} & \multicolumn{2}{c}{1e-3} \\
                \textbf{$R_{cb}$} & \multicolumn{2}{c}{4} \\
                \textbf{$F_1$} & \multicolumn{2}{c}{0.33} \\
                \textbf{$F_2$} & \multicolumn{2}{c}{0.5} \\
                \textbf{$R$} & 1000 & 2000 \\
                \textbf{$K$} & 64 & 128 \\
                \textbf{$Dir(\beta)$} & 10.0 &  0.1 \\
                \bottomrule \newline
            \end{tabular}%
        }
        \caption{$\rho$=$0.1$.\label{tab:mobilenet_hp_lp}}
    \end{subtable}
\end{table*}

\begin{table*}[!h]
    \small \centering
    \caption{\small{Hyperparameters used with YAMNet architecture.}\label{tab:yamnet_hp}}
    \begin{subtable}[t]{0.21\textwidth}
        \resizebox{1.0\textwidth}{!}{%
            \begin{tabular}{cc}
                \toprule
                \textbf{Parameter} & \begin{tabular}[c]{c}\textbf{\textit{IID}}\\\textbf{($C_p\approx1.0$)}\end{tabular}\\
                \midrule
                \textbf{\textit{params}} & 3.21M \\
                \textbf{$N$} & 10 \\
                \textbf{$E$} & 4 \\
                \textbf{$\rho$} & 1.0 \\
                \textbf{$lr$} & 1e-3 \\
                \textbf{$R_{cb}$} & 4 \\
                \textbf{$F_1$} & 0.33 \\
                \textbf{$F_2$} & 0.5 \\
                \textbf{$R$} & 100 \\
                \textbf{$K$} & 64 \\
                \textbf{$Dir(\beta)$} & 10.0 \\
                \bottomrule \newline
            \end{tabular}%
        }
        \caption{$\rho$=$1.0$\label{tab:yamnet_hp_mp}}
    \end{subtable}
    \hspace{1.0cm} %
    \begin{subtable}[t]{0.21\textwidth}
        \resizebox{1.0\textwidth}{!}{%
            \begin{tabular}{cc}
                \toprule
                \textbf{Parameter} & \begin{tabular}[c]{c}\textbf{\textit{IID}}\\\textbf{($C_p\approx1.0$)}\end{tabular}\\
                \midrule
                \textbf{\textit{params}} & 3.21M \\
                \textbf{$N$} & 10 \\
                \textbf{$E$} & 4 \\
                \textbf{$\rho$} & 0.1 \\
                \textbf{$lr$} & 1e-3 \\
                \textbf{$R_{cb}$} & 4 \\
                \textbf{$F_1$} & 0.33 \\
                \textbf{$F_2$} & 0.5 \\
                \textbf{$R$} & 1000 \\
                \textbf{$K$} & 64 \\
                \textbf{$Dir(\beta)$} & 10.0 \\
                \bottomrule \newline
            \end{tabular}%
        }
        \caption{$\rho$=$0.1$.\label{tab:yamnet_hp_lp}}
    \end{subtable}
\end{table*}

\section{Accuracy - Communication Cost Trade-off\label{sec:appendix_d}}

\begin{figure*}[!h]
    \centering \small
    \captionsetup[subfigure]{justification=centering}
    \begin{subfigure}[b]{0.45\textwidth}
        \centering
        \includegraphics[width=\linewidth]{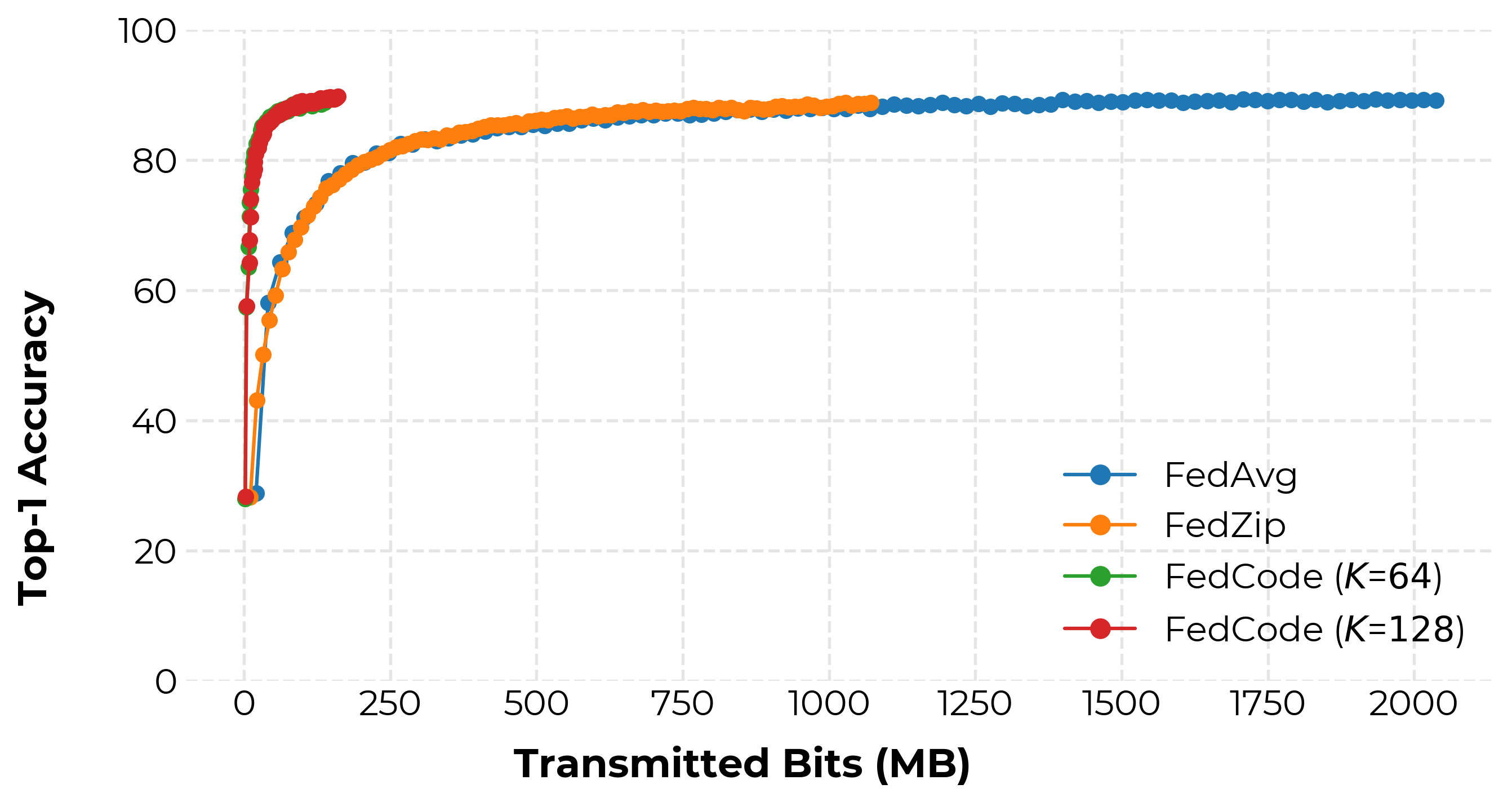}
        \vspace{0.1cm}
        \caption{\small{CIFAR-10 on IID settings}\label{fig:dtr_acc_iid_linear}}
    \end{subfigure}
    \hspace{0.5cm}
    \begin{subfigure}[b]{0.45\textwidth}
        \centering
        \includegraphics[width=\linewidth]{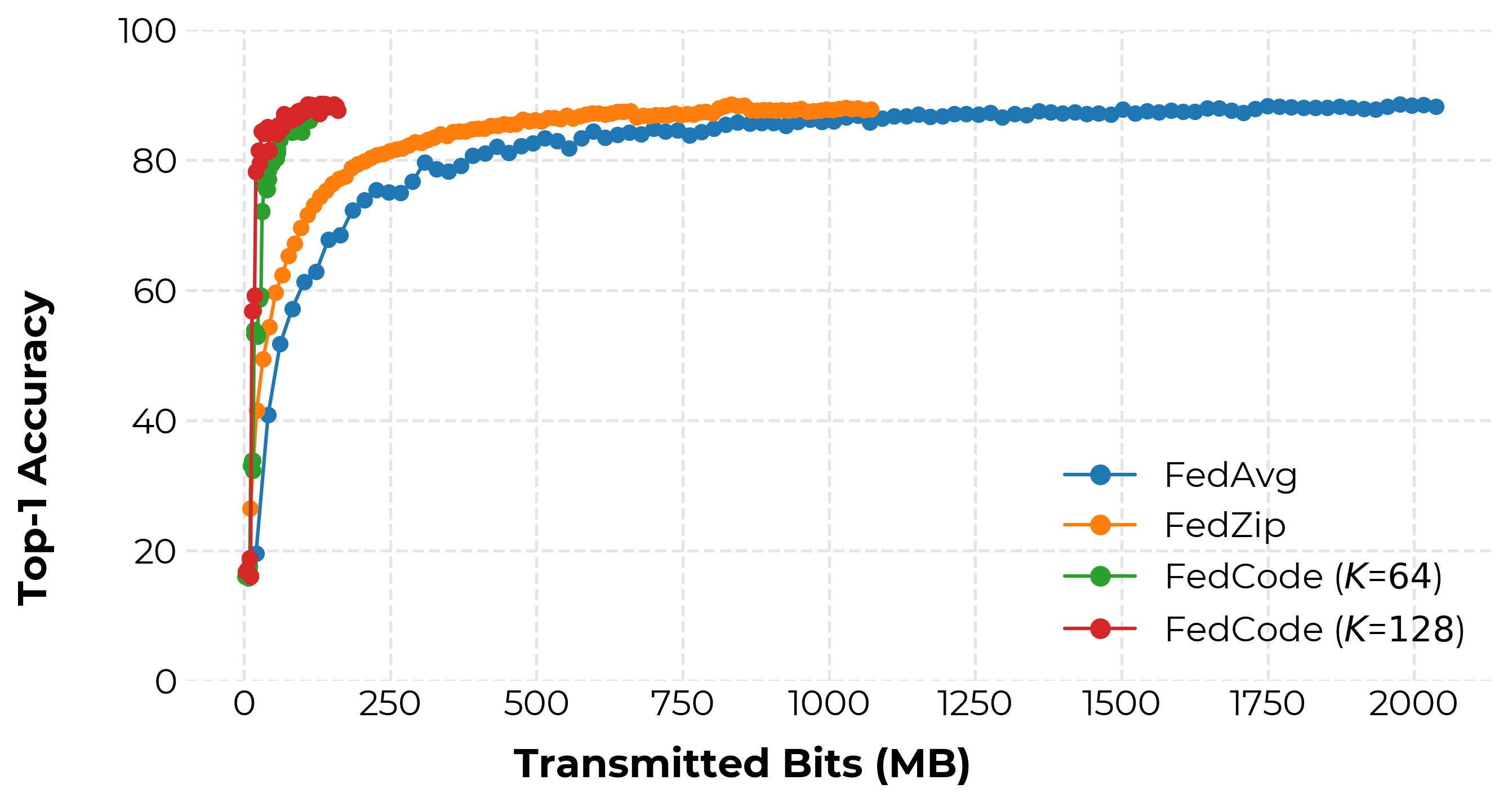}
        \vspace{0.1cm}
        \caption{\small{CIFAR-10 on non-IID settings}\label{fig:dtr_acc_noniid_linear}}
    \end{subfigure}
    \caption{\small{Accuracy versus Volume of Communicated Data trade-off for~\method~using ResNet-20 architecture in CIFAR-10 for (a) IID, and (b) non-IID settings. In both figures, the transmitted bits are reported in linear scale.}\label{fig:dtr_acc_linear}}
\end{figure*}

To vividly illustrate the benefits of~\method~in reducing the volume of communicated data, we present results analogous to those in Figure~\ref{fig:dtr_acc}. In this version, the volume of communicated data is projected in terms of MBs transmitted, displayed in a linear scale. As evident from Figure~\ref{fig:dtr_acc_linear}, significant communication savings can be achieved by frequently leveraging the codebook transfer during the FL training stage.

\section{Bitrate Considerations\label{sec:appendix_e}}

Bandwidth efficiency is paramount in the context of FL, especially in environments where network resources are scarce or expensive. As the number of edge devices participating in FL grows, ensuring efficient utilization of the available bandwidth becomes critical. In this section, we consider the bits-per-parameter ($bpp$) required across FL training stage to clearly illustrate the benefits of~\method~in terms of bandwidth saving. Specifically, a low $bpp$ can lead to faster transmission times, reduced network congestion, and minimized communication overhead, enhancing both FL scalability and ensuring seamless collaboration among devices with diverse communication capabilities. From Figure~\ref{fig:bpp}, we notice that~\method~ can reduce the average $bpp$ to below 1-bit threshold, achieving similar results to complex quantization schemes, such as~\cite{eden, fedpm} with minimal computation overhead.

\begin{figure*}[!h]
    \centering \small
    \captionsetup[subfigure]{justification=centering}
    \begin{subfigure}[b]{0.45\textwidth}
        \centering
        \includegraphics[width=\linewidth]{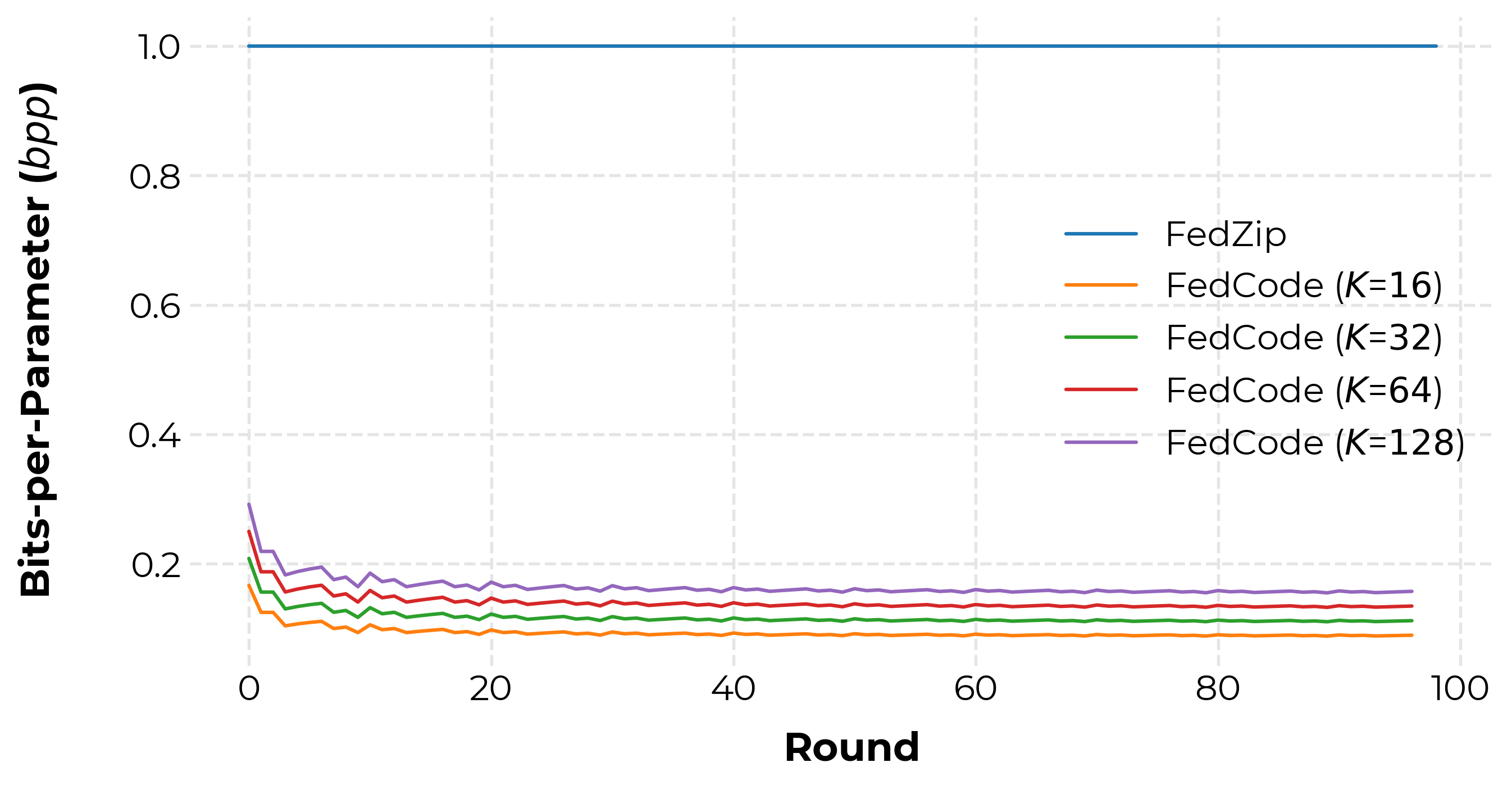}
        \vspace{0.1cm}
        \caption{\small{Bitrate Performance of~\method.}\label{fig:bpp}}
    \end{subfigure}
    \hspace{0.5cm}
    \begin{subfigure}[b]{0.45\textwidth}
        \centering
        \includegraphics[width=\linewidth]{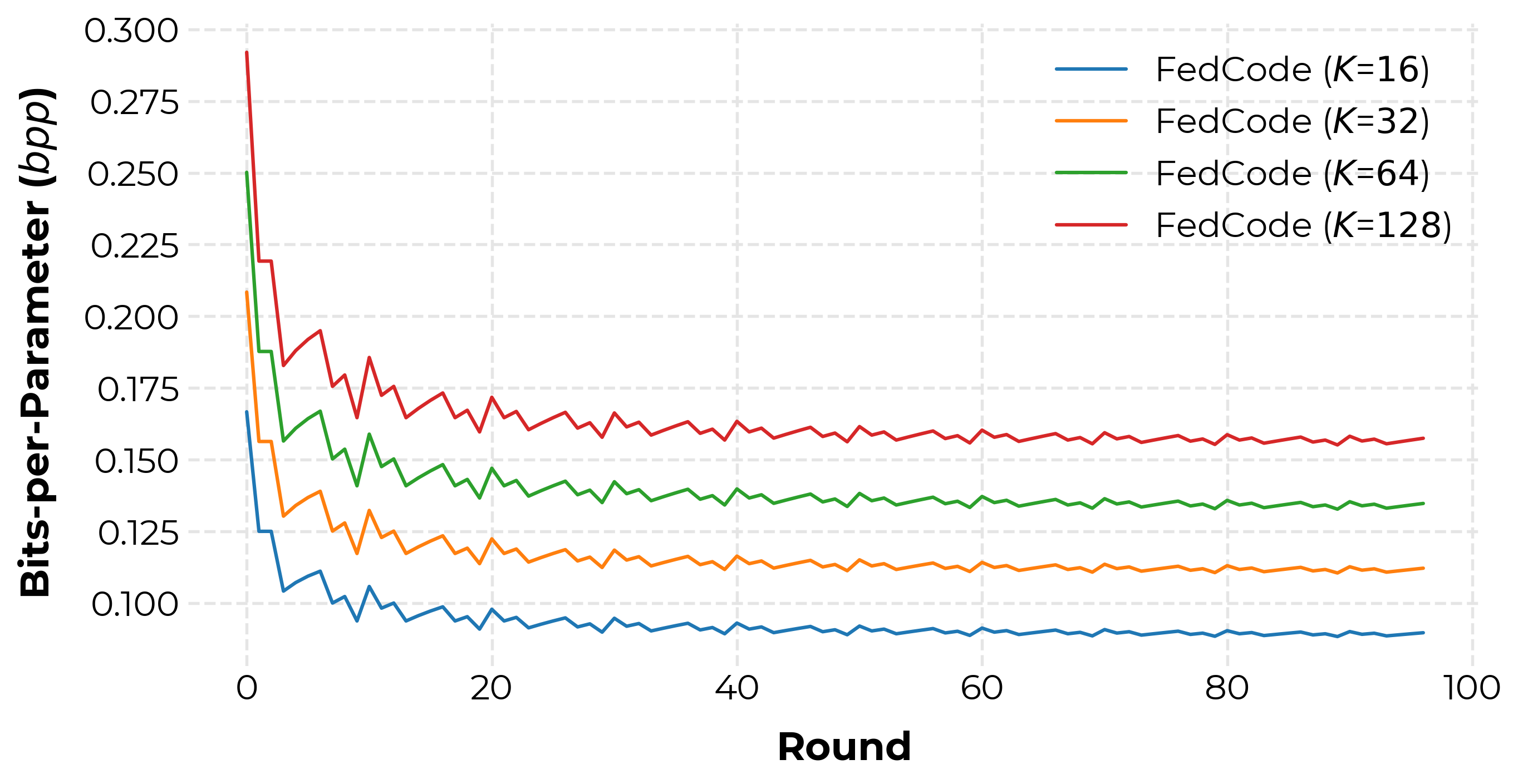}
        \vspace{0.1cm}
        \caption{\small{Zoomed view of (a) in lower bitrates.}\label{fig:bpp_zoom}}
    \end{subfigure}
    \caption{\small{Average bits-per-parameters required from~\method~using ResNet-20 architecture in CIFAR-10 for IID settings with $F_1$=$0.2$ and $F_2$=$0.5$.}\label{fig:bpp}}
\end{figure*}

\end{document}

%% file: methodology.tex
We introduce~\method, which primarily relies on transferring codebooks in the communication between server and clients in FL, allowing for significant reductions in the volume of transmitted data during training and utilizing minimal bandwidth in the communication channel. As cluster centers can shift across federated rounds, we periodically transmit the compressed weight matrices (weight values replaced with cluster indexes) to ensure consistency across the models of clients and the server. Thus,~\method~comprises two types of training rounds that alternate. The first type is those rounds in which weight matrices are updated solely by updating the codebook, increasing communication-efficiency. The second type is those rounds in which codebook updates and the compressed weights matrices are both communicated.

\subsection{\textbf{Federated Learning with Transferring Codebook}}
An overview of the FL process relying solely on codebook transmission is depicted in Figure \ref{fig:overview}. Since we transfer the codebook instead of the actual weight matrices, both local model update and server-side model aggregation phases require adjustments from standard FL process. In what follows, we describe these processes in detail for both downstream (server to clients) and upstream (clients to server) communication routes.

\begin{itemize}

\item \textit{Downstream Communication Route:} On the server side, we extract the global model's weights codebook by applying weight clustering using K-means~\cite{kmeans} clustering algorithm, which we sort and broadcast to all participating clients of the current federated round. Once the codebook is received on a client, a \textit{local weights update} step is performed to update the locally stored model. This process consists of a computationally inexpensive binary search on the sorted codebook, where each weight value of the local model's weight matrices is substituted with the nearest center in the codebook. Once the locally stored model weights have been updated, the clients can proceed with their local training step. 

\item \textit{Upstream Communication Route}: 
Once clients perform their local training step on their locally stored data, weight clustering will be performed on the new weight matrices, the generated codebook of which will be transmitted back to the server. With the collection of clients' codebook being completed at the server side, a \textit{global weights update} step will be performed, which involves concatenation of received codebooks and updating the global model's weight values with their nearest center in the ``\textit{aggregated}'' codebook. From here, we apply a clustering algorithm, such as K-means~\cite{kmeans}, on the global model's weights and create the new codebook to be utilized in the next federated round.

\end{itemize}

\subsection{\textbf{Cluster Calibration}}

When clustering the model's weights into clusters, we assign a single value (i.e, a center) to represent the weight values within each cluster. The concept of transferring the codebook is based on the idea that the weight values within each cluster only change slightly between subsequent federated rounds. Therefore, instead of updating each individual value, we can update the center of each cluster. However, it is important to consider that the distribution of weight values (i.e., clusters) can change, meaning that the position of weight values within the clusters may shift after training the clients' models or updating the global model. Updating weight values with the center of a cluster to which they no longer belong can result in model performance degradation. As a result, transferring the codebook for a large number of rounds without updating the clusters may prevent the global model from converging and achieving higher performance. To tackle this problem, it is necessary to calibrate both the global model, client models, and the clusters periodically. 

To calibrate the models, we introduce a cluster calibration phase, during which we transfer the compressed weights along with their codebook. Once clients receive the compressed weight matrices and corresponding codebook, they decompress the global model's weight matrices by replacing each index in the compressed weight matrices with its respective codebook center. Alternatively, if the server receives the compressed weights, it decompresses them in a similar fashion, after which global model weights aggregation is performed using \textit{FedAvg}~\cite{fedavg}. The step-by-step procedure for implementing~\method~can be found in Algorithm \ref{alg:method}, whereas Algorithm~\ref{alg:client_algo} provides the details of the client-side operation.

\input{algorithm}

\input{algorithm_client_process}

\subsection*{\textbf{Volume of Transmitted Data}}

In~\method, we exchange the codebook between the clients and the server at each federated round. To ensure clusters are calibrated correctly, we periodically transmit the compressed weight matrices in addition to the codebook. Revisiting Equation~\ref{eqn:dtr}, in~\method~the volume of exchanged messages in terms of bits downloaded and uploaded, is as follows:

\begin{equation}\label{eq:}
    \resizebox{0.8\hsize}{!}{%
        $ R \times \left( 2 \times K \times \textit{worldlength}  + (F_1 + F_2) \times \left(  P \times \lceil \log_2 K \rceil \right ) \right ) $,
    }
\end{equation}

\noindent where $R$ represents the number of rounds, $P$ denotes the number of trainable parameters, and $K$ is the number of clusters utilized by the clustering algorithm. With $F_1$ and $F_2$, we express the frequency of transmission (scalars in range of $[0,1]$) of the compressed weight matrices for the downstream and upstream communication, respectively. We provide a theoretical analysis of potential reductions in data transmission achieved by~\method~in Appendix~\ref{sc:dtr_anlysis}.

\subsection*{\textbf{Client-side Computational Overhead}}
Since clients are often resource-constrained devices, it is essential for our approach to introduce a low computation overhead for the clients. On the downstream communication route, clients receive a sorted codebook and execute a \textit{local weights update} step, during which model weights are replaced with their nearest cluster centers using a simple binary search on the sorted array of centers. Hence, the computational complexity of this downstream process is:

\begin{equation}
    \mathcal{O}(P \cdot \log K),
\end{equation}

\noindent where $P$ represents the number of trainable parameters in the clients' models, and $K$ is the number of clusters.

Alternatively, during upstream communication, clients employ a clustering algorithm to extract cluster centers and generate their codebook. For K-means~\cite{kmeans} clustering, the computational complexity of this task is:

\begin{equation}
    \mathcal{O}(P \cdot K \cdot I),
\end{equation}

\noindent where $I$ represents the number of iterations in the K-means~\cite{kmeans} algorithm.

%% file: algorithm.tex
\begin{algorithm}[!t]
    \centering \small
    \caption{\small{\method: Federated learning with transferring Codebooks. In the algorithm, \texttt{cluster} refers to K-means clustering with $C$ number of clusters. Scalar $R_{cb}$ indicates the activation round for codebook transfer process, while $F_1$ and $F_2$ are the communication frequency of compressed weights for downstream and upstream communication channel, respectively.\label{alg:method}}}
    \begin{algorithmic}[1]
        \State Server initialization of model with model weights $\theta^{0}$
        \For{ $i=1, \dots, R$ }
            \State Randomly select $M$ clients to participate in round $i$
            \For{ each client $m \in M$ \textbf{ in parallel}}
                \State $\left(\mathcal{C}^{i},\theta^{i} \right)$ $\gets$ \texttt{cluster} $\left ( \theta^{i} \right )$
                \If{$\left(i \mod \left(1/F_1\right) \right)$ = $0$ \textbf{and} $r > R_{cb}$ }
                    \State $z_{m}^{i+1}$ $\gets$ $\textbf{ClientUpdate}$($\left(\mathcal{C}^{i},\theta^{i}\right )$,$i$)
                \Else
                    \State $z_{m}^{i+1}$ $\gets$ $\textbf{ClientUpdate}$(\texttt{sort}$\left ( \mathcal{C}^{i} \right )$,$i$)
                \EndIf
            \EndFor
            \If{$\left(i \mod \left(1/F_2\right) \right)$ = $0$ \textbf{and} $r > R_{cb}$}
                \State $\left (\mathcal{C}, \theta \right)$ $\gets$ $z_{m}^{i+1}$
                \State $\theta_m^{i+1}$  $\gets$ $\texttt{decompress} \left (\mathcal{C}, \theta \right)$
                \State $\theta^{i+1} \gets \sum\nolimits_{m=1}^{M} \frac{N_k}{N}~~\theta_m^{i+1}$
            \Else
                \State $\mathcal{C}^{i+1} \gets \texttt{sort} \left(\texttt{concat}\left (\sum\nolimits_{m=1}^{M} z_m^{i+1} \right) \right)$
                \For{$t \in \theta^{i}$}
                    \State $t \gets \texttt{binary\_search} \left(\mathcal{C}^{i+1},t \right)$
                \EndFor
            \EndIf
        \EndFor
    \end{algorithmic}
\end{algorithm}

%% file: algorithm_client_process.tex
\begin{algorithm}[!t]
    \centering \small
    \caption{\small{Update process at the client side. Here, $\mathcal{L}$ is the loss function, $p_{\theta}$ refers to a network with parameters $\theta$. With $\eta$ we refer to the learning rate, while $F_1$ and $F_2$ are the communication frequency of compressed weights for downstream and upstream communication channel, respectively.\label{alg:client_algo}}}
    \begin{algorithmic}[1]
        \Procedure{ClientUpdate}{$x$,$r$}
            \If{$\left(r \mod \left(1/F_1\right) \right)$ $\neq$ $0$ \textbf{and} $r > R_{cb}$}
                \State $\mathcal{C} \gets x $
                \For{$t \in \theta^{r-1}$}
                    \State $t \gets \texttt{binary\_search} \left ( \mathcal{C}, t \right)$
                \EndFor
            \Else
                \State $\left (\mathcal{C}, \theta \right)  \gets x $
                \State $\theta$ $\gets$ $\texttt{decompress} \left (\mathcal{C}, \theta \right)$
            \EndIf
            \For{epoch $e=1,2,\dots,E_{c}$}
                \For{ batch $b \in \mathcal{D}_{l}$}
                    \State $\theta \gets \theta - \eta_{c} \cdot \nabla_{\theta} \left(\mathcal{L} \left( p_{\theta}\left(b\right)\right)\right)$
                \EndFor
            \EndFor
            \State $\left(\mathcal{C},\theta \right)$ $\gets$ \texttt{cluster}$\left ( \theta \right )$ 
            \If{$\left(r \mod \left(1/F_2\right) \right)$ = $0$ \textbf{and} $r > R_{cb}$}
                \Return $\left ( \mathcal{C} , \theta \right) $
            \EndIf
            \State \Return $\mathcal{C}$
        \EndProcedure
    \end{algorithmic}
\end{algorithm}